\pdfoutput=1

\documentclass[11pt]{article}

\usepackage{emnlp2022}

\usepackage[T1]{fontenc}

\usepackage{times}
\usepackage{latexsym}
\usepackage{url}

\usepackage{microtype}

\usepackage[export]{adjustbox}
\usepackage{amsmath}
\usepackage{amssymb}
\usepackage{bbm}
\usepackage{boldline}
\usepackage{booktabs}
\usepackage{enumitem}
\usepackage{hyperref}
\usepackage{multirow}
\usepackage{caption}
\usepackage{subcaption}
\usepackage{tabularx}
\usepackage{tikz}
\usepackage{xspace}
\usepackage{xargs}
\usepackage[colorinlistoftodos,prependcaption,textsize=tiny]{todonotes}
\usetikzlibrary{positioning}
\usepackage{pifont}

\newcommand{\set}[1]{\mathcal{#1}}

\usepackage{makecell}

\newcommand{\cut}[1]{}

\newcommand\tf[1]{\textbf{#1}}

\newcommand\red[1]{\textcolor{red}{#1}}

\title{Improving Passage Retrieval with Zero-Shot Question Generation}

\author{
Devendra Singh Sachan$^{1,2}$\Thanks{\hspace{.03in}This work was done during an internship at Meta AI.}, Mike Lewis$^{3}$, Mandar Joshi$^{4}$, Armen Aghajanyan$^{3}$, \\
{\bf Wen-tau Yih$^{3}$, Joelle Pineau$^{1,2,3}$, Luke Zettlemoyer$^{3,5}$} \\
$^{1}$McGill University; 
$^{2}$Mila - Quebec AI Institute \\
$^{3}$Meta AI Research; $^{4}$Google Research; $^{5}$University of Washington \\
{\tt sachande@mila.quebec, mandarj@google.com}\\
\tt \{mikelewis,armenag,scottyih,jpineau,lsz\}@meta.com\\
}

\begin{document}

\maketitle


\begin{abstract}
We propose a simple and effective re-ranking method for improving passage retrieval in open question answering.
The re-ranker re-scores retrieved passages with a  zero-shot question generation model, which uses a pre-trained language model to compute the probability of the input question conditioned on a retrieved passage. 
This approach can be applied on top of any retrieval method (e.g.\ neural or keyword-based), does not require any domain- or task-specific training (and therefore is expected to generalize better to data distribution shifts), and provides rich cross-attention between query and passage (i.e. it must explain every token in the question). 
When evaluated on a number of open-domain retrieval datasets, our re-ranker improves strong unsupervised retrieval models by 6\%-18\% absolute and strong supervised models by up to 12\% in terms of top-20 passage retrieval accuracy.
We also obtain new state-of-the-art results on full open-domain question answering by simply adding the new re-ranker to existing models with no further changes.\footnote{Our codebase including data and checkpoints is available at: \url{https://github.com/DevSinghSachan/unsupervised-passage-reranking}}

\end{abstract}


\section{Introduction}
\label{sec:introduction}

Text retrieval is a core sub-task in many NLP problems, for example, open-domain question answering where a document must be retrieved and then read to answer an input query. Queries and documents are typically embedded in a shared representation space to enable efficient search, before using a task-specific model to perform a deeper, token-level document analysis (e.g.\ a document reader that selects an answer span). We show that adding a zero-shot re-ranker to the retrieval stage of such models
leads to large gains in performance, by doing deep token-level analysis with no task-specific data or tuning. 

\begin{figure}[t!]
\centering
\includegraphics[max width=\linewidth, scale=1.0]{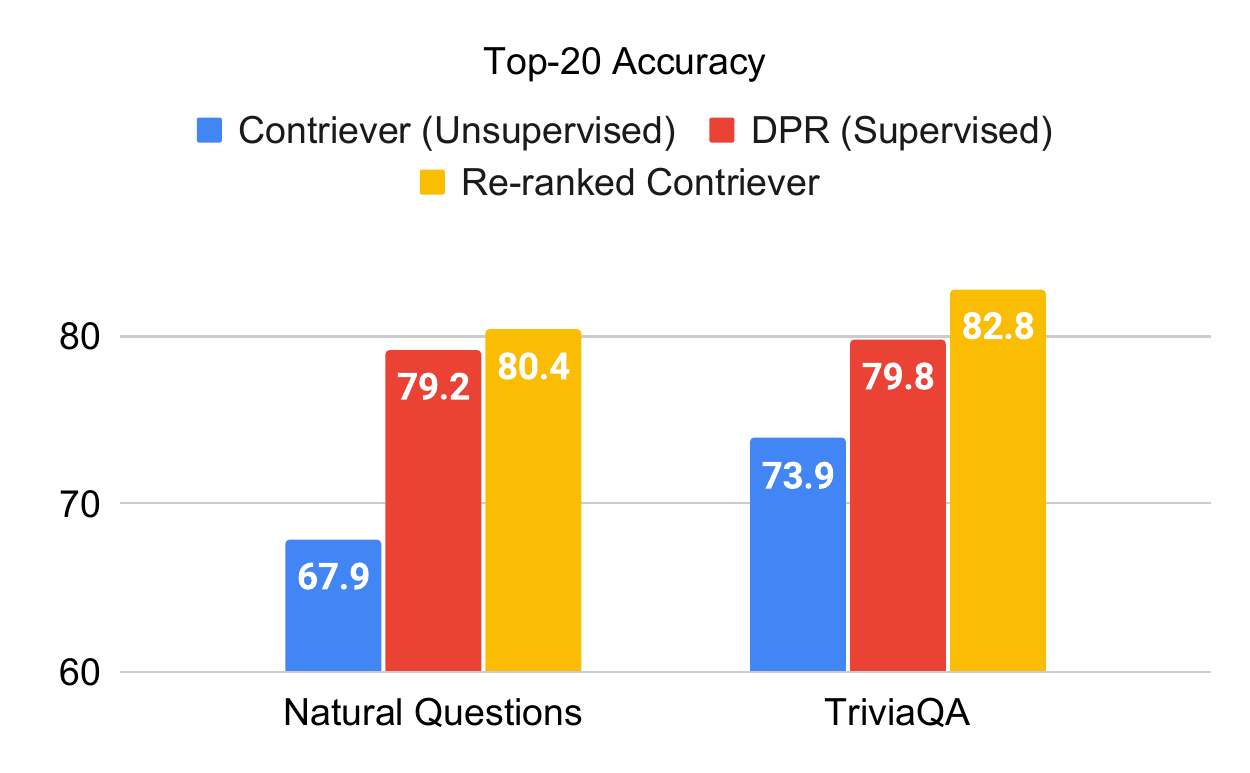}
\caption{
After UPR re-ranking of the  Contriever's (unsupervised)~\cite{izacard2021towards} top-1000 passages, we outperform strong supervised models like DPR~\cite{karpukhin2020dense} on Natural Questions and TriviaQA datasets.
}
\label{fig:intro-figure}
\end{figure}

We focus on open-domain question answering and introduce a re-ranker based on zero-shot question generation with a pre-trained language model. 
Our re-ranker, which we call \emph{Unsupervised Passage Re-ranker} (UPR), re-scores the retrieved passages by computing the likelihood of the input question conditioned on a retrieved passage.\footnote{In this paper, we refer to the words documents and passages interchangeably. We consider the retrieval units as short passages and not entire documents.}
This simple method enables task-independent cross-attention between query and passage that can be applied on top of any retrieval method (e.g.\ neural or keyword-based) and is highly effective in practice (Figure~\ref{fig:intro-figure}). 


In part, UPR is inspired by the traditional models of query scoring with count-based language models~\cite{10.1145/383952.384019}.
However, instead of estimating a language model from each passage, UPR uses pre-trained language models (PLMs).
More recent work on re-rankers have finetuned PLMs on question-passage pairs to generate relevance labels~\cite{nogueira-etal-2020-document}, sometimes to jointly generate question and relevance labels~\cite{nogueira-dos-santos-etal-2020-beyond,ju2021text}.
In contrast, UPR uses off-the-shelf PLMs, does not require any training data or finetuning, and still leads to strong performance gains (Figure~\ref{fig:intro-figure}).

Comprehensive experiments across a wide range of datasets, retrievers, and PLMs highlight the strengths of UPR:

\begin{itemize}[noitemsep,leftmargin=*]
	\item By re-ranking the top-1000 passages from Contriever (unsupervised), UPR obtains a gain of 6\%-18\% points absolute in the top-20 retrieval accuracy across four QA datasets. UPR also achieves new state-of-the-art results on the difficult SQuAD-Open and Entity Questions datasets, outperforming BM25 by 14\% and 8\%.

	\item These performance gains are consistent across both different kinds of retrievers and PLMs. Ablation studies reveal that instruction-tuned models such as T0 perform the best as re-rankers. 

    \item On the open-domain QA task, just by performing inference with the re-ranked passages and a pre-trained reader, we obtain improvements of up to 3 EM points on three benchmark datasets.
\end{itemize}

To the best of our knowledge, this is the first work to show that a fully unsupervised pipeline (consisting of a retriever and re-ranker) can greatly outperform supervised dense retrieval models like DPR~\cite{karpukhin2020dense}.
As language models continue to improve rapidly \cite{Rae2021ScalingLM, Chowdhery2022PaLMSL}, the performance of UPR may see corresponding gains over time.
UPR requires no annotated data and uses only generic pre-trained models, which means it may be easy to apply to a wide range of retrieval problems.

\cut{
The rest of the paper is organized as follows. 
Sec.~\ref{sec:method} details the unsupervised re-ranking method. 
Sec.~\ref{sec:exp-setup} and ~\ref{sec:exp-passage-retrieval} describe the experimental settings and the results, respectively. 
Sec.~\ref{sec:related-work} reviews the related work followed by discussion in Sec.~\ref{sec:conclusion}.
}


\section{Method}
\label{sec:method}


\begin{figure*}[ht!]
\centering
\includegraphics[max width=\textwidth, scale=0.85]{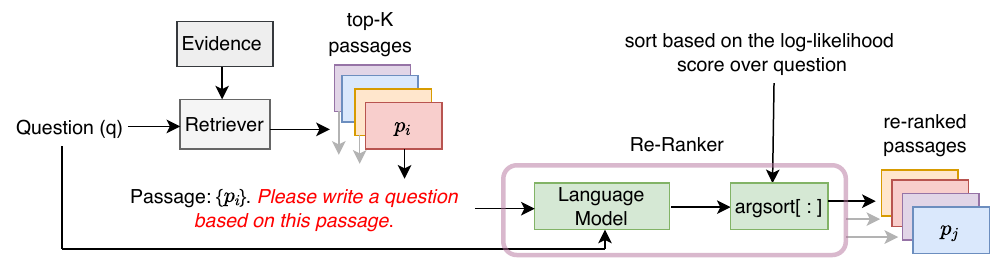}
\caption{An illustration of the different components in UPR. For more details, please refer to text.}
\label{fig:reranking-method}
\end{figure*}

Figure~\ref{fig:reranking-method} presents an overview of our approach for open-domain retrieval, which introduces a new unsupervised re-ranker (Sec~\ref{sec:method-upr}) that can be applied to any existing text retriever (Sec~\ref{sec:method-retriever}). 

\subsection{Retriever}
\label{sec:method-retriever}

Let $\set{D} = \{\boldsymbol{d}_1, \ldots, \boldsymbol{d}_M \}$ be a collection of evidence documents.
Given a question ($\boldsymbol{q}$), the retriever selects a subset of relevant passages $\set{Z} \subset \set{D}$, one or more of which will ideally contain the answer to $\boldsymbol{q}$.
Our method will work with passages obtained from any retriever --- either based on sparse representations like BM25 or dense representations like DPR.
We only assume that the retriever provides the $K$ most relevant passages. 
We denote this set of \emph{top-K} passages as $\mathcal{Z} = \{\boldsymbol{z}_1, \ldots, \boldsymbol{z}_K\}$. 

\subsection{Unsupervised Passage Re-ranking (UPR)}
\label{sec:method-upr}

Given the top-K retrieved passages, the goal of the re-ranker is to reorder them such that a passage with the correct answer is ranked as highly as possible.
The ordering is computed with a \emph{relevance score} $p(\boldsymbol{z}_i \mid \boldsymbol{q})$ for each passage $\boldsymbol{z}_i \in \set{Z}$.

Our re-ranking approach is unsupervised, \emph{i.e.}, it does not use any task-specific training examples. We refer to it as \textbf{UPR}, for \emph{Unsupervised Passage Re-ranking}.
UPR uses a pre-trained language model to score the probability of generating the question $\boldsymbol{q}$ given the passage text $\boldsymbol{z}$, as described below. The question generation model is zero-shot, allowing for  dataset-independent re-ranking, and also incorporates cross-attention between the question and passage tokens while forcing the model to explain every token in the input question. UPR is, therefore, more expressive than using dense retrievers alone, even if both methods fundamentally build on top of the same (or very similar) pre-trained models.

More specifically, we estimate $p(\boldsymbol{z}_i \mid \boldsymbol{q})$ by computing the likelihood of \emph{question generation} conditioned on the passage, \emph{i.e.}, the quantity $p(\boldsymbol{q} \mid \boldsymbol{z}_i)$.
This also naturally emerges when applying Bayes' rule to $p(\boldsymbol{z}_i \mid \boldsymbol{q})$ as
\begin{align*}
\log p(\boldsymbol{z}_i \mid \boldsymbol{q}) = \log p(\boldsymbol{q} \mid \boldsymbol{z}_i) + \log p(\boldsymbol{z}_i) + c\ ,
\end{align*}
where $p(\boldsymbol{z}_i)$ is the prior on the retrieved passage and $c$ is a common constant for all $\boldsymbol{z}_i$. 

As a simplifying assumption, we assume that the passage prior $\log p(\boldsymbol{z}_i)$ is uniform, and can be ignored for re-ranking. 
With this, the above expression reduces to
\begin{align*}
\log p(\boldsymbol{z}_i \mid \boldsymbol{q}) \propto \log p(\boldsymbol{q} \mid \boldsymbol{z}_i),\ \forall \boldsymbol{z}_i \in \set{Z}\ . 
\end{align*}

We estimate $\log p(\boldsymbol{q} \mid \boldsymbol{z}_i)$ using a pre-trained language model (PLM) to compute the average log-likelihood of the question tokens conditioned on the passage: 
\begin{align*}
\log p(\boldsymbol{q} \mid \boldsymbol{z}_i) = \frac{1}{|\boldsymbol{q}|}\sum_t \log p(q_t \mid \boldsymbol{q}_{<t}, \boldsymbol{z}_i; \Theta)\ .
\end{align*}
where $\Theta$ denotes the parameters of the PLM and $|\boldsymbol{q}|$ denotes the number of question tokens. We apply the PLM in a zero-shot fashion with no finetuning by simply appending the natural language instruction ``\emph{Please write a question based on this passage}'' to the passage tokens as shown in Figure~\ref{fig:reranking-method}.

The initial passage ordering is then sorted based on $\log p(\boldsymbol{q} \mid \boldsymbol{z})$.
This enables us to re-rank the passages by just performing inference using off-the-shelf language models avoiding the need to label question-passage pairs for finetuning. Because the question generation model is applied zero-shot, this overall approach can be applied to improve the retrieval accuracy of any test collection, with no dataset-specific models or tuning data.


\section{Experimental Setup}  \label{sec:exp-setup}
In this section, we describe the datasets, unsupervised and supervised retrievers, and language models used for our passage re-ranking experiments. 

\subsection{Open-Domain QA Datasets} \label{subsec:datasets}

Following previous work on passage retrieval, we use the popular datasets of  SQuAD-Open~\cite{rajpurkar-etal-2016-squad}, TriviaQA~\cite{joshi2017triviaqa}, Natural Questions (NQ;~\citet{Kwiatkowski2019natural}), and WebQuestions (WebQ;~\citet{berant-etal-2013-semantic}). 
Their statistics are presented in Table~\ref{tab:dataset_stats}.

\paragraph{Evidence Passages $\mathcal{D}$.} We use the preprocessed English Wikipedia dump from December 2018 as released by \citet{karpukhin2020dense} as our evidence passages. Each Wikipedia article is split into non-overlapping 100 word passages. There are over 21 million total passages.


\begin{table}[t]
\centering
\small
\begin{tabular}{@{}l |c c c c@{}}
 \toprule
 \textbf{Dataset} & \textbf{\ Train} & \textbf{Ret.\ Train}  & \textbf{Dev} & \textbf{Test} \\
 \midrule
 SQuAD-Open             & 78,713 & 70,096 & 8,886 & 10,570\\
 TriviaQA          & 78,785 & 60,413 & 8,837 & 11,313 \\
 NQ                & 79,168 & 58,880 & 8,757 & \phantom{0}3,610 \\
 WebQ              & \phantom{0}3,417 & \phantom{0}2,474 & \phantom{00}361 &  \phantom{0}2,032 \\
 \bottomrule
 \end{tabular}
\caption{QA dataset statistics. Full training set examples are used for the open-domain QA experiments, while the retriever training examples are used for supervised retriever training. We leverage these dataset splits as provided by~\cite{karpukhin2020dense}.
}
\label{tab:dataset_stats}
\end{table}

\subsection{Keyword-centric Datasets}
To examine the robustness of UPR to keyword-centric datasets, we experiment with test collections where dense retrievers struggle and when the questions are from different domains.

\paragraph{Entity Questions} contains 22K short questions about named entities based on facts from Wikipedia. 
Previous work on this dataset has shown that dense retrievers struggle to retrieve relevant passages while sparse approaches like BM25 are more successful~\cite{sciavolino2021simple}.

\paragraph{BEIR Benchmark} is a test suite for benchmarking retrieval algorithms and consists of multiple datasets, where each dataset consists of test set queries, evidence documents, and relevance document annotations~\cite{thakur2021beir}.
These datasets contain different kinds of retrieval tasks like fact-checking, question answering, etc.\ and span diverse domains including news, technical, and Wikipedia making it a challenging benchmark.

\subsection{Retrievers}  \label{subsec:retrievers}
In our re-ranking experiments, we retrieve passages from both unsupervised and supervised retrievers, as detailed below.

\subsubsection{Unsupervised Retrievers}

\paragraph{BM25} ranks based on the term-frequency and inverse document frequency of the keywords present in the question and passage~\cite{Robertson2009bm25}. 
Prior work~\cite{ma2021replication} has shown that BM25 is a strong baseline for the datasets we consider.

\paragraph{MSS} is a dense retriever trained by predicting masked salient spans like named entities with the help of a reader network~\cite{sachan2021end}. 
MSS pre-training has also shown to improve supervised retrieval performance.

\paragraph{Contriever} uses momentum contrastive training to learn dense retrievers from text paragraphs~\cite{izacard2021towards}. 
Such training has shown to obtain strong zero-shot retrieval performance on many benchmarks.

\subsubsection{Supervised Retrievers}

\paragraph{DPR} uses annotated question-context paragraphs and hard negative examples to train a supervised dense retriever~\cite{karpukhin2020dense}.

\paragraph{MSS-DPR} further improves DPR performance by first pre-training the dense retriever using MSS followed by DPR-style supervised finetuning~\cite{sachan2021end}.

\subsection{Pre-Trained Language Models (PLMs)}  \label{subsec:plms}

We use a range of pre-trained models for computing our re-ranking relevance scores.  

\paragraph{T5 Series} These models consist of encoder and decoder transformers pre-trained by denoising input text sequences. We experiment with the T5 model~\cite{raffel2020t5}, its language model adapted version (T5-lm-adapt;~\citet{lester-etal-2021-power}), and the T0 language model~\cite{sanh2022multitask}.
T0 was trained by finetuning T5-lm-adapt with multiple tasks defined by instructions. 
We use the ``xl'' configurations that contain 3B parameters.

\paragraph{GPT} These consist of a transformer decoder trained with the autoregressive language modeling objective.
We use the GPT-neo model with 2.7B parameters~\cite{gpt-neo}.

\subsection{Implementation Details}
\label{subsec:implementation_details}

We run all the experiments on a cluster with V100-32GB GPUs. 
We use PyTorch~\cite{paszke2019pytorch} to implement the UPR approach and relevant baselines.
To get the top-K retrieved passages, we use the open-source implementations of the retrievers and their checkpoints.
For BM25, we use the pre-computed top-k passages outputs from the pyserini toolkit~\cite{Lin_etal_SIGIR2021_Pyserini}.\footnote{\url{https://github.com/castorini/pyserini/blob/master/docs/experiments-dpr.md}}
For MSS, DPR, and MSS-DPR retrievers, we use the open-source implementations from~\cite{sachan2021endtoend}.\footnote{\url{https://github.com/DevSinghSachan/emdr2}}
For Contriever and PLMs, we use their checkpoints as hosted in Huggingface~\cite{wolf-etal-2020-transformers}.

For the dense retriever experiments, we use the \emph{base configuration}, which consists of 12 attention heads, 12 layers, and 768 model dimensions. 
To experiment with supervised retrievers, we train DPR and MSS-DPR for 3 epochs on SQuAD-Open, 40 epochs on NQ and TriviaQA, and 20 epochs on WebQ.\footnote{In contrast to previous work on SQuAD-Open, we train DPR and MSS-DPR for 3 epochs to prevent overfitting.}
Detailed hyperparameter settings are specified in Appendix~\ref{appendix:train-hparams} and \ref{appendix:instruction-prompt}.


\section{Experiments: Passage Retrieval}
\label{sec:exp-passage-retrieval}

\begin{table*}[!ht]
\addtolength{\tabcolsep}{-0.65pt}
\small
\centering
\begin{tabular}{@{}l | c c | c c | c c | c c | c c@{}}
\toprule
\tf{Retriever} & \multicolumn{2}{c}{\tf{SQuAD-Open}} & \multicolumn{2}{c}{\tf{TriviaQA}} & \multicolumn{2}{c}{\tf{NQ}} & \multicolumn{2}{c}{\tf{WebQ}} & \multicolumn{2}{c}{\tf{Average}} \\
                    & Top-20 & Top-100 & Top-20 & Top-100 & Top-20 & Top-100 & Top-20 & Top-100 & Top-20 & Top-100 \\
\midrule
\multicolumn{11}{c}{\textit{Unsupervised Retrievers}} \\
\midrule
MSS         &       51.3  &   68.4  &  67.2  &   79.1  &   60.0 &  75.6  &   49.2  & 68.4 &  56.9 & 72.9 \\
MSS + UPR         &   75.7  &   80.8  &  81.3  &   85.0  &   77.3 &  81.5  &   71.8  & 80.4  & 76.5 & 81.9 \\
\midrule
BM25        &       71.1  &   81.8  &  76.4  &   83.2  &   62.9 &  78.3   &  62.4  & 75.5 &  68.2 & 79.7 \\
BM25 + UPR        &  \underline{83.6} & \underline{87.4} &  \underline{83.0}  &   \underline{86.4}  &   78.6 &  85.2   &  72.9  & 81.4 & 79.5 & 85.1 \\
\midrule
Contriever & 63.4  &   78.2  &  73.9  &   82.9  &   67.9 &  80.6   &  65.7  & 80.1 &  70.0 & 80.5 \\
Contriever + UPR  &   81.3  &   85.6  &  82.8  &   \underline{86.4}  &   \underline{80.4} &  \underline{87.0}   &  \underline{75.7}  & \underline{83.5} & \underline{80.1} & \underline{85.6} \\
\midrule
\multicolumn{11}{c}{\textit{Supervised Retrievers}} \\
\midrule
DPR        &        59.4  &   74.5  &  79.8  &   85.1  &   79.2 &  85.7   &  74.6  & 81.6 &  73.3 & 81.7 \\
DPR + UPR &   80.7  &   85.4  &  84.3  &   87.2  &   83.4 &  88.6   &  77.7  & 84.1  &  81.5 & 86.3 \\
\midrule
MSS-DPR    &        73.1  & 84.5    &  81.9  &   86.6  &   81.4 &  88.1   &  76.9  & 84.6 &  78.3 & 86.0 \\
MSS-DPR + UPR     &   \tf{85.2}&\tf{89.4}&\tf{84.8}&\tf{88.0}& \tf{83.9} & \tf{89.4} & \tf{77.2} & \tf{85.2} & \tf{82.8} & \tf{88.0} \\
\midrule
E2E Supervised &  -   &  -   & 84.1   &   87.8  &   84.8 & 89.8 &  79.1  &  85.2  \\
\bottomrule
\end{tabular}
\caption{
Top-\{20, 100\} retrieval accuracy on the test set of datasets before and after UPR re-ranking of the top-1000 retrieved passages with the T0-3B model. 
Best results of the unsupervised retriever are underlined while those of the supervised retriever are highlighted in bold.
For reference, we also include the state-of-the art supervised results in the last row, which is obtained from end-to-end or joint training of the retriever and language model using question-answer pairs~\cite{sachan2021end,sachan2021endtoend}.
}
\label{tab:qa-reranking}
\end{table*}

We evaluate the performance of our proposed Unsupervised Passage Re-ranker (UPR), conduct ablations to better understand the approach, evaluate robustness on challenging test collections, and discuss run-time efficiency.

Our goal is to improve the rankings of top-\{20, 100\} passages. 
Hence, in the first stage, a larger candidate list is fetched by retrieving the top-1000 passages.
Then, in the second stage, these passages are re-ranked with the T0-3B PLM unless specified otherwise.
To evaluate UPR performance, we compute the conventional \emph{top-K retrieval accuracy} metric. It is defined as the fraction of questions for which at least one passage within the top-K passages contains a span that matches the human-annotated answer to the question.

\subsection{Main Task} 
\label{subsec:main-task}

We experiment with the four datasets and five retrievers as introduced in \S\ref{subsec:datasets} and \S\ref{subsec:retrievers}, respectively and 
perform re-ranking with the T0-3B model.
Table~\ref{tab:qa-reranking} reports the top-20 and top-100 retrieval accuracy before and after re-ranking.
UPR provides consistent improvements across all the retrievers and datasets, improving unsupervised models by 6\%-18\% absolute and supervised models by up to 12\% in top-20 accuracy.

Re-ranked Contriever outperforms DPR by an average of 7\% in top-20 and 4\% in top-100 when considering all the datasets.
This shows that \emph{a fully unsupervised pipeline of a retriever and re-ranker can outperform strong supervised models like DPR}.
Sparse representations still remain competitive, with BM25 outperforming Contriever and MSS on SQuAD-Open and TriviaQA re-ranking.

We also see that re-ranked MSS-DPR comes close to or matches the performance of state-of-the-art supervised retrievers (last row in Table~\ref{tab:qa-reranking}).
Because these supervised models are based on end-to-end training of the retriever and language model, they are memory-intensive and too expensive to train for very large models.
As such, \emph{UPR offers a viable alternative to expensive joint training}.

\paragraph{Intuition behind the performance gains obtained by UPR}
The question generation step in the re-ranker involves expressive cross-attention with the passage tokens. As a result, each question token attends to all the passage tokens in each decoder layer before predicting the next question token. This results in an accurate estimation of the relevance (or log-likelihood) scores than the original retriever scores, thus leading to an improved retrieval accuracy after re-ranking. 
This reasoning is further corroborated by our error analysis in Appendix~\ref{subsec:analysis}, where we present several examples where UPR improves over the incorrect BM25 retrievals.

\subsection{Ablation Studies} 

\subsubsection{Importance of Question Generation}
\label{why-quesgen}


\begin{figure}[t]
\centering
\includegraphics[max width=\linewidth, scale=1.0]{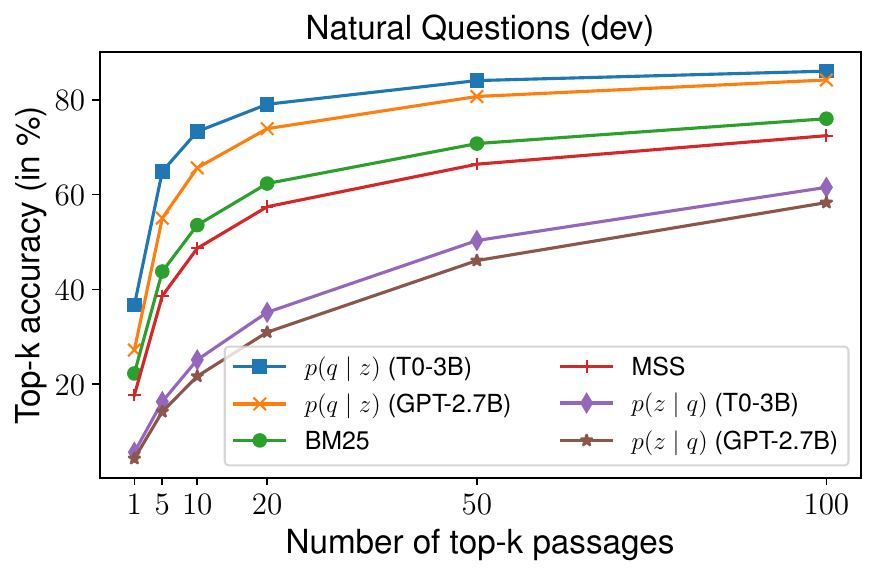}
\caption{
Comparison of two passage re-ranking approaches on the NQ development set:
(1) when generating question tokens conditioned on the passage $p(\boldsymbol{q} \mid \boldsymbol{z})$, and
(2) when generating passage tokens conditioned on the question $p(\boldsymbol{z}\mid \boldsymbol{q})$.
Results highlight the usefulness of question generation in UPR for re-ranking.
}
\label{fig:why-quesgen}
\end{figure}

To understand the importance of re-ranking based on question generation $p(\boldsymbol{q}\mid \boldsymbol{z})$, we compare it with another unsupervised approach where re-ranking is based on passage generation conditioned on the question $p(\boldsymbol{z}\mid \boldsymbol{q})$.
This quantity can be estimated by computing the average log-likelihood of generating the passage tokens using PLM and teacher-forcing as
\begin{align*}
\log p(\boldsymbol{z} \mid \boldsymbol{q}; \Theta) = \frac{1}{|\boldsymbol{z}|}\sum_t \log p(z_t \mid \boldsymbol{z}_{<t}, \boldsymbol{q}; \Theta)\ ,
\end{align*}
where $\Theta$ denotes the parameters of the PLM and $|\boldsymbol{z}|$ denotes the number of passage tokens.

For this analysis, we work with the NQ development set and obtain the union of top-1000 passages from the BM25 and MSS retrievers. These passages are re-ranked with two PLMs: T0-3B and GPT-2.7B. 
Our results in Figure~\ref{fig:why-quesgen} demonstrate that question generation obtains substantial improvements over the BM25 and MSS, highlighting its usefulness in passage re-ranking.
On the other hand, re-ranking based on passage generation leads to a drop in retrieval accuracy in comparison to the baseline retrievers, empirically confirming that this approach does not work well in practice.

\subsubsection{Impact of Pre-trained Language Models}




\begin{table}[t]
\small
\centering
\begin{tabular}{@{}l | c c c c @{}}
 \toprule
 \tf{Retriever /} & \multicolumn{4}{c}{\tf{NQ (dev)}}  \\
 \tf{Re-Ranker}                   &  Top-1 & Top-5 & Top-20 & Top-100  \\
\midrule
       BM25       & 22.3  & 43.8  & 62.3   &  76.0  \\  
       MSS        & 17.7  & 38.6  & 57.4   &  72.4 \\
 \cmidrule{2-5}
T5 (3B)  &          22.0  & 50.5  & 71.4   &  84.0 \\
GPT-neo (2.7B)      & 27.2  & 55.0  & 73.9   &  84.2 \\
GPT-j (6B)      & 29.8 & 59.5 & 76.8 & 85.6 \\
T5-lm-adapt (250M) & 23.9 & 51.4 & 70.7 & 83.1 \\
T5-lm-adapt (800M) &  29.1 & 57.5 & 75.1 & 84.8 \\
T5-lm-adapt (3B) &  29.7  & 59.9  & 76.9   & 85.6 \\
T5-lm-adapt (11B) &  32.1 & 62.3 & 78.5 & 85.8 \\
T0-3B            & 36.7 & \tf{64.9} & \tf{79.1} & \tf{86.1} \\
T0-11B           & \tf{37.4} & \tf{64.9} & \tf{79.1} & 86.0 \\

\bottomrule
\end{tabular}
\caption{
Comparison of different pre-trained language models (PLMs) as re-rankers on the NQ development set.
We re-rank the union of BM25 + MSS retrieved passages with UPR. 
Results demonstrate that T0 PLMs achieves the best top-K accuracy among the compared PLMs.
}
\label{tab:qa-reranking-plms}
\end{table}

To understand how much the choice of PLM contributes to top-K accuracy, we compare the performance of T5 (3B), T5-lm-adapt (different sizes), T0-\{3B, 11B\}, and GPT-neo (2.7 B) (as introduced in \S\ref{subsec:plms}) on the NQ development set.
We obtain the union of top-1000 passages retrieved from BM25 and MSS and then re-rank them with UPR. 
Results in Table~\ref{tab:qa-reranking-plms} reflect that all the PLMs obtain significant improvements over the baseline retrievers, with the T0 models achieving the best results. 
Scaling up the PLM size, especially the T5-lm-adapt models, leads to consistent performance improvements.

When comparing across PLMs, we see that the performance of T5 suffers especially on top-\{1, 5\} accuracy levels. This might be because it was trained to predict corrupted spans, which is not ideal for text generation.
On the other hand, autoregressive PLMs such as GPT-neo and T5-lm-adapt tend to be better re-rankers.
Furthermore, T0 obtains large improvements on top-\{1, 5, 20\}, demonstrating that finetuning with instructions on unrelated tasks is also beneficial for re-ranking.

\subsubsection{Passage Candidate Size vs Latency}
\label{subsec:latency}


\begin{figure}[t!]
\centering
\includegraphics[max width=\linewidth, scale=1.0]{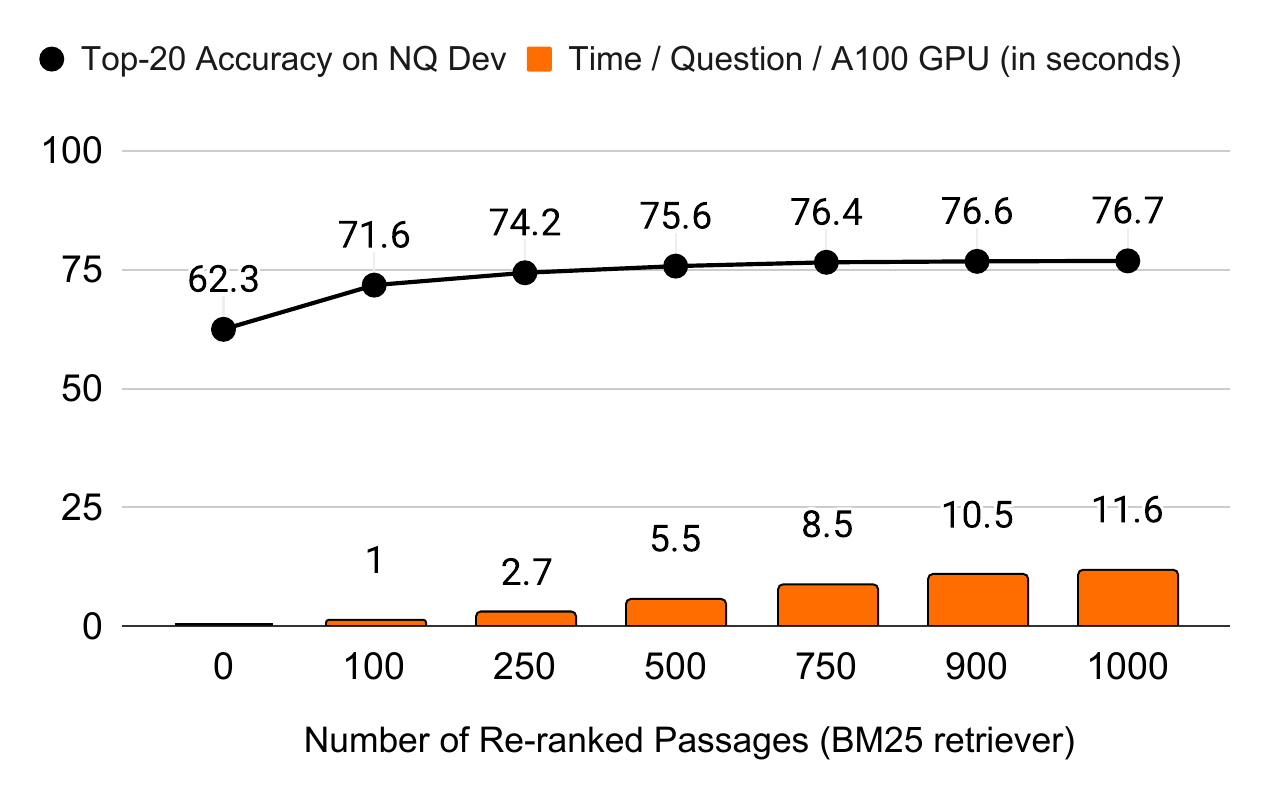}
\caption{
Effect of the number of passage candidates on top-20 accuracy and latency when re-ranked with T0-3B PLM. Evaluation is done on the NQ development set using BM25 retrieved passages.
}
\label{fig:runtime-efficiency}
\end{figure}

We study the effect of the number of passage candidates to be re-ranked on the retrieval performance along with the time taken.
For this, we consider the NQ development set, re-rank up to top-1000 passages obtained from BM25, and use top-20 accuracy as the evaluation criteria.
Results in Figure~\ref{fig:runtime-efficiency} illustrate that a larger pool of passage candidates indeed helps to improve the performance.
However, the gains tend to plateau as the number of passages is increased.

With more passages, the latency in re-ranking per question linearly increases reflecting the trade-off between accuracy and throughput. 
The higher latency can be partly alleviated with approaches like weight quantization, efficient implementations of the transformer kernel, model distillation, caching passage embeddings, and using data parallelism.
However, we leave these explorations to future work. 

\subsection{Zero-Shot Supervised Transfer}
To gain a better understanding of the relative strengths of UPR and supervised (or finetuned) re-rankers, we perform zero-shot supervised transfer experiments and compare the results with UPR.
We adopt the training method of ~\citet{nogueira-etal-2020-document}, henceforth referred to as \emph{monoT5}, who finetune
the T5 PLMs on the MS MARCO~\cite{bajaj2016ms} passage ranking dataset. To train, question and passage tokens are concatenated and fed to the T5 encoder. The decoder attends to the encoded sequence and the T5 PLM is finetuned to maximize the likelihood of the ``true'' label. To re-rank the passages during inference, the log-likelihood score of the ``true'' label is used as the relevance score.

We use the open-source checkpoints of monoT5 to re-rank the top-1000 passages retrieved by BM25 and report results on the NQ development set (Table~\ref{tab:supervised-transfer}).\footnote{\url{https://github.com/castorini/pygaggle}}
Interestingly, we see that supervised transfer improves the top-1 and top-5 retrieval accuracy by a large margin over UPR. 
However, when the set of retrieved passages increases, such as 20-100, the results of UPR come close to or match the results of monoT5.
As end tasks such as open-domain question answering rely on a larger set of passages to achieve good results (as demonstrated in Sec~\ref{sec:question-answering}), \emph{this highlights the importance of UPR over supervised models as it does not require collecting annotated data for finetuning}.
\begin{table}[t]
\small
\centering
\begin{tabular}{@{}l | c c c c @{}}
 \toprule
 \tf{Retriever /} & \multicolumn{4}{c}{\tf{NQ (dev)}}  \\
 \tf{Re-Ranker}                   &  Top-1 & Top-5 & Top-20 & Top-100  \\
\midrule
       BM25       & 22.3  & 43.8  & 62.3   &  76.0  \\ 
 \cmidrule{2-5}
UPR (T0-3B) & 36.1 & 62.8 & 76.8 & 83.1 \\
monoT5 (250M) & 39.1 & 62.4 & 75.6 & 82.6 \\
monoT5 (800M) & 43.5 & 66.1 & 77.5 & 83.3 \\
monoT5 (3B) & \textbf{44.2} & \textbf{68.3} & \textbf{78.7} & \textbf{83.7} \\
\bottomrule
\end{tabular}
\caption{
Zero-shot supervised transfer results on the NQ development set.
We use monoT5~\cite{nogueira-etal-2020-document} checkpoints of different sizes finetuned on the MS MARCO dataset to re-rank the top-1000 passages retrieved by BM25. 
We also include the results of UPR for reference.
}
\label{tab:supervised-transfer}
\end{table}

\subsection{Evaluation on Keyword-centric Datasets}
\label{subsec:keyword-centric}

\subsubsection{Entity Questions}
\label{result:entity-questions}

We re-rank the top-1000 passages from every retriever with UPR.
As the training set is not provided, we use the checkpoints of DPR and MSS-DPR trained on NQ.
Results are presented in Table~\ref{tab:entity-questions}.
Re-ranking leads to a gain of 8-20\% absolute in top-20 accuracy and 4-10\% in top-100 accuracy, with BM25 achieving the best results after re-ranking.
It also substantially narrows the gap between BM25 and dense retrievers.
Re-ranking the union of BM25 and Contriever outputs outperforms the current best results by 6\% and 3\% in top-20 and top-100, respectively.

We also note that multi-vector approaches specially tailored towards the robust representation of textual entities~\cite{jong2022mention} are promising alternatives to the dual-encoder retrievers as they offer an improved retrieval accuracy although at the expense of increased memory and compute requirements. However, we defer the application of UPR to these retrievers as a part of future work.

\subsubsection{BEIR Benchmark}      \label{results:beir-benchmark}


\begin{table}[!t]
\small
\centering
\begin{tabular}{l | c c}
 \toprule
 \tf{Retriever} & \multicolumn{2}{c}{\tf{Entity Questions}}  \\
                & Top-20 & Top-100  \\
\midrule
\multicolumn{3}{c}{\textit{Baselines}} \\
\midrule
MSS  & 51.2 & 66.3 \\
DPR   &  51.1 &  63.8 \\
MSS-DPR     & 60.6 & 73.7 \\
Contriever  & 63.0 & 75.1 \\
BM25       &       71.2 & 79.8 \\ 
SPAR~\cite{chen2021salient} & 74.0 & 82.0 \\
\midrule
\multicolumn{3}{c}{\textit{After Re-ranking with UPR (T0-3B PLM)}} \\
\midrule
MSS        & 71.3 & 76.7 \\
DPR        & 65.4 & 72.0   \\ 
MSS-DPR    & 73.9 & 80.1 \\
Contriever & 76.0 & 81.6 \\
BM25       & 79.3 & 83.9 \\ 
BM25 + Contriever& \tf{80.2} & \tf{85.4} \\
\bottomrule
\end{tabular}
\caption{Top-\{20, 100\} retrieval accuracy on the Entity Questions dataset before and after re-ranking.
Following the original paper, we report macro-average scores.
}
\label{tab:entity-questions}
\end{table}


\begin{table}[t]
\small
\centering
\begin{tabular}{l | c c}
 \toprule
 \tf{Retriever} & \multicolumn{2}{c}{\tf{BEIR}}  \\
                & nDCG@10 & Recall@100  \\
\midrule
\multicolumn{3}{c}{\textit{Baselines}} \\
\midrule
BERT~\cite{devlin2019bert} & \phantom{0}9.3 & 20.1 \\
SimCSE~\cite{gao-etal-2021-simcse} & 27.4 & 48.1 \\
REALM~\cite{guu2020realm} & 25.8 & 46.5 \\
Contriever  &  36.0   & 60.1 \\
BM25       &  41.6 &  63.6\\ 
\midrule
\multicolumn{3}{c}{\textit{After Re-ranking with UPR (T0-3B PLM)}} \\
\midrule
Contriever &  44.6 & 66.3 \\
BM25       &  \tf{44.9} & \tf{68.0} \\ 
\bottomrule
\end{tabular}
\caption{Macro-average nDCG@10 and Recall@100 scores on the BEIR benchmark.
Performance numbers of the baseline models are from~\citet{izacard2021towards}.
}
\label{tab:beir-summary}
\end{table}

We re-rank the top-1000 documents from Contriever and BM25 with the T0-3B PLM.
Following convention, we report the macro average of NDCG@10 and Recall@100 metrics in Table~\ref{tab:beir-summary}. 
Results demonstrate the effectiveness of UPR as NDCG@10 scores improve by 3-8\% absolute and Recall@100 improves by 5-6\%.
We include performance numbers on individual datasets with fine-grained analysis in Appendix~\ref{appendix:beir-benchmark}.

\section{Experiments: Question Answering}
\label{sec:question-answering}


\begin{table*}[!ht]
\small
\centering
\begin{tabular}{l c c c c c c c c}
 \toprule
 \textbf{Model} & top-$K$ & \multicolumn{2}{c}{\textbf{SQuAD-Open}} & \multicolumn{2}{c}{\textbf{TriviaQA}} & \multicolumn{2}{c}{\textbf{NQ}} & \textit{\# of} \\
                &    &  dev & test & dev & test & dev & test & \textit{params}\\
 \midrule
 \multicolumn{9}{c}{\textbf{Baselines}} \\
\midrule
 BM25 + BERT~\cite{lee-etal-2019-latent}  & \phantom{00}5     & 28.1 & 33.2 & 47.2 & 47.1 & 24.8 & 26.5 & 220M \\
 ORQA~\cite{lee-etal-2019-latent}         & \phantom{00}5     & 26.5 & 20.0 & 45.1 & 45.0 & 31.3 & 33.3 & 330M \\
 REALM~\cite{guu2020realm}                & \phantom{00}5     &  -   &  -   &  -   &  -   & 38.2 & 40.4 & 330M \\
 DPR~\cite{karpukhin2020dense}            & \phantom{0}25     &  -   & 38.1 &  -   & 56.8 &  -   & 41.5 & 330M \\
 RAG-Sequence~\cite{lewis20rag}           & \phantom{0}50     &  -   &   -  & 55.8 & 56.8 & 44.0 & 44.5 & 626M \\
 Individual Top-K (large)~\cite{sachan2021end} &  -           &  -   &   -  &  -   & 59.6 &  -   & 48.1 & \phantom{0}1.2B \\
 Joint Top-K (large)~\cite{sachan2021end}   & \phantom{0}50   &  -   &   -  &  -   & 68.3 &  -   & 51.4 & \phantom{0}1.2B \\
 FiD-base~\cite{izacard2021leveraging}    & 100               &  -   & 53.4 &  -   & 65.0 & -    & 48.2 & 440M \\
 FiD-large~\cite{izacard2021leveraging}   & 100               &  -   & 56.7 &  -   & 67.6 & -    & 51.4 & 950M \\
 FiD-KD-base~\cite{izacard2021distilling} & 100               &  -   &  -   & 68.6 & 68.8 & 48.0 & 49.6 & 440M \\
 FiD-KD-large~\cite{izacard2021distilling}& 100               &  -   &  -   & 71.9 & 72.1 & 51.9 & 53.7 & 950M \\
 $\text{EMDR}^2$-base~\cite{sachan2021endtoend}& \phantom{0}50 &  46.8 & 51.1 & 71.1 & 71.4 & 50.4 & 52.5 & 440M \\
\midrule
\multicolumn{9}{c}{\textbf{Our Implementation}} \\
\midrule
FiD-base (MSS retriever, T5 reader) & \multirow{2}{*}{100}     & 36.2 & 39.6  &  60.9 & 60.3  & 43.7 & 44.5 &   \multirow{2}{*}{440M} \\
\phantom{0000000}+ Inference with UPR re-ranked passages  &                & 43.7 & 50.1  &  68.5 & 68.9  & 45.8 & 47.3 & \\
\midrule
FiD-base (DPR retriever, T5 reader) & \multirow{2}{*}{100}     &    48.8 & 45.8  &  67.9 & 68.5 & 49.4 & 50.8 & \multirow{2}{*}{440M} \\
\phantom{0000000}+ Inference with UPR re-ranked passages &                  &   51.5 & 54.0  &  70.1 & 71.2 & 49.8 & 51.3 \\
\midrule
FiD-base (MSS-DPR retriever, T5 reader) & \multirow{2}{*}{100} & 50.1   & 52.2 & 69.9 & 70.2 & 49.7 & 50.8    & \multirow{2}{*}{440M} \\
\phantom{0000000}+ Inference with UPR re-ranked passages   &               & 51.9   & 55.6 & 71.5 & 71.8 & 49.9 & 51.5 \\
\midrule
FiD-large (MSS-DPR retriever, T5 reader)& \multirow{2}{*}{100} & 51.9  & 54.4 &  71.5    & 71.6 & \tf{51.8} & 53.6 & \multirow{2}{*}{950M} \\
\phantom{0000000}+ Inference with UPR re-ranked passages   &    & \tf{53.1}  & \tf{58.1} & \tf{72.7} & \tf{73.2} & 51.5 & \tf{54.5} \\
 \bottomrule
\end{tabular}
\caption{Exact match scores for the open-domain QA task. We train one FiD model for each retriever as indicated and then perform inference with its respective re-ranked outputs. Top-$K$ denotes the number of retrieved passages that are used by the reader to produce an answer. 
We report the baseline performance numbers from the respective papers.
The best performing models are highlighted in bold.
}
\label{tab:odqa-ansgen}
\end{table*}

Finally, we show that UPR improves the performance of full open-domain QA systems. 

\subsection{Method}
An open-domain QA system consists of a retriever and a reader component.
The reader attends to the retrieved passages to produce a final answer to the question. 
We use the Fusion-in-Decoder (FiD;~\citet{izacard2021leveraging}) model as the reader.
In FiD, each retrieved passage is concatenated with the question and is then passed as an input to the T5 encoder~\cite{raffel2020t5}. Then the encoded representations for all the passages are concatenated which the T5 decoder leverages for cross-attention.

We train the FiD reader using standard negative log-likelihood loss and teacher-forcing to generate an answer autoregressively. 
To understand the effect of UPR on answer generation, we then do inference with the previously trained reader and the re-ranked passages for each question.

\subsection{Results}
For training FiD models, we use the top-100 retrieved passages and a batch size of 64.
Detailed training hyperparameters are provided in Appendix~\ref{appendix:train-hparams}.
During inference, an answer is generated using greedy decoding.
For our experiments, we train the FiD base and large models using the retrieved documents from MSS, DPR, and MSS-DPR retrievers. 
We re-rank the top-1000 passages with UPR using the T0-3B PLM and then perform inference with the top-100 re-ranked passages.
We conduct experiments on SQuAD-Open, TriviaQA, and NQ datasets and report the exact match (EM) scores for evaluation. 
We employ the same set of evidence passages for all the datasets.\footnote{Previous work has often used the 2016 Wikipedia dump as evidence for SQuAD-Open. As our evidence set is larger and newer, some questions may be unanswerable, which renders a fair comparison difficult. However, to alleviate dataset-specific design choices, we adopt a common experimental setup.}

Results are presented in Table~\ref{tab:odqa-ansgen}. 
More accurate passages after re-ranking improve the performance of the pre-trained FiD models for all the retrievers.
Performing inference on the FiD-large model with re-ranked MSS-DPR passages achieves new state-of-the-art results, outperforming the pre-trained FiD model by 1-3 EM points.
Overall, this provides a simple approach for obtaining performance gains without the need to iteratively re-train~\cite{izacard2021distilling} or perform expensive end-to-end training~\cite{sachan2021endtoend}.


\section{Related Work}
\label{sec:related-work}

Our work is based on re-ranking passages for open-domain retrieval using pre-trained language models (PLMs) which we have covered in earlier sections.
Here, we instead focus on covering previous work related to generative pre-training, query likelihood for document ranking, and open-domain QA.

\paragraph{Generative Pre-training and Instruction Tuning}
Recently, there has been an increased adoption of the generative pre-trained transformer (GPT) series of models by the NLP community~\cite{radford2019language}.
Among the interesting properties of GPT models is their ability to understand task instructions specified in natural language and then perform well on tasks in a zero-shot or few-shot manner~\cite{NEURIPS2020_1457c0d6,smith2022using}.
The zero-shot performance of GPT models further improves when finetuning them on multiple different tasks using task-specific instructions, which is also known as instruction-tuning~\cite{sanh2022multitask,wei2022finetuned,min2021metaicl}.

\paragraph{Document Ranking based on Query Likelihood}
In information retrieval, an appealing approach to rank documents is by utilizing language models to compute relevance scores for a query~\cite{10.1145/290941.291008}.
Prior approaches estimated a count-based language model for each document that was used to compute query likelihood scores for ranking~\cite{10.1145/383952.384019}.
However, these approaches suffer from issues such as data sparsity. 
More recent approaches utilize PLMs such as GPT or T5 to compute query likelihood~\cite{nogueira-dos-santos-etal-2020-beyond}. 
To improve ranking accuracy, they perform supervised finetuning using query-document pairs~\cite{ju2021text}.
Our work also utilizes PLMs, but instead, we leverage a larger instruction-tuned language model and apply them in a zero-shot manner without finetuning.

\paragraph{Open-Domain QA}
involves producing answers to information-seeking questions from large document collections.
Typical approaches consist of retriever and reader networks, where the retriever identifies a small number of documents to aid the reader in producing answers~\cite{chen2017reading}.
To be scalable, retrievers are often modeled using dual-encoders~\cite{lee-etal-2019-latent} or with multi-vector encoders~\cite{zhou-devlin-2021-multi} and then to further improve retieval accuracy, re-rankers are employed~\cite{nogueira-etal-2020-document}.
Given retrieved documents, a reader is then trained to generate a short answer to the question~\cite{izacard2021leveraging,sachan2021endtoend}.


\section{Conclusions and Future Work}
\label{sec:conclusion}

In this work, we propose UPR, an approach to perform unsupervised passage re-ranking for open-domain retrieval.
To re-rank, UPR computes a relevance score for question generation conditioned on each retrieved passage using pre-trained language models. 
Extensive experiments across a wide range of QA datasets show that an unsupervised pipeline consisting of retriever and UPR greatly outperforms strong supervised retriever models.
In addition, UPR further improves the performance of supervised retrievers. 
On the open-domain QA task, by just performing inference using re-ranked passages and a pre-trained reader model, we achieve new state-of-the-art results.

UPR presents several interesting directions for future work.
First, its applications to other retrieval tasks such as improving source-code retrieval based on textual queries can be explored.
Second, another promising direction would be to tune instructions according to the nature of the retrieval tasks. 
For instance, when retrieving similar sentences in the BEIR benchmark, variations of the instruction prompt used in UPR can be explored.
Finally, it would also be interesting to investigate the extent to which specialized language models such as the ones finetuned to generate questions using passage-questions data  would further help in improving retrieval.



\section*{Acknowledgements}
This work was done during the first author's internship at Meta AI Research.
The authors would like to thank Dmytro Okhonko and the anonymous reviewers for providing useful suggestions and feedback about this work that helped us to improve the paper.
We would also like to thank the administrators of the compute cluster at FAIR, Meta AI for their assistance in facilitating experimental runs.

\section*{Limitations}
A limitation of UPR is that re-ranking a large pool of passages can have a high latency as it involves performing cross-attention whose complexity is proportional to the product of the question and passage tokens and the number of layers of the pre-trained language model (PLM).
We have also discussed this quantitatively in Sec \ref{subsec:latency}.
UPR also shares the inherent limitation associated with all the re-ranking approaches in that its maximum possible performance is dependent on the first-stage retrieval. For example, when processing the top-1000 retrieved passages, the upper limit of top-100 re-ranking accuracy would be the top-1000 accuracy of the retrieved passages.
Finally, we want to remark that UPR results might be sensitive to the training data used to train the PLM. As a result, in a domain-specific retrieval or question-answering task, PLMs trained on in-domain text~\cite{gururangan2020don} are expected to be more accurate than those trained on broad-coverage text.

\section*{Ethics Statement}
The experiments conducted in the paper demonstrate the usefulness of large language models for information retrieval tasks when using English Wikipedia as the evidence source. However, when deployed in production, our work shares the typical ethical risks associated with large language models.
There are chances that the re-ranked results may not be fair to all communities. This can potentially lead to an increased discrimination and exclusion of marginalized groups. These risks can also perpetuate to question-answering applications such as generating toxic or fake text as answers. Therefore, care should be taken before deploying our approach in real-world or customer facing applications; it is advisable to conduct tests and benchmark the models covering these aspects.

\bibliography{custom}

\begin{thebibliography}{45}
\expandafter\ifx\csname natexlab\endcsname\relax\def\natexlab#1{#1}\fi

\bibitem[{Bajaj et~al.(2016)Bajaj, Campos, Craswell, Deng, Gao, Liu, Majumder,
  McNamara, Mitra, Nguyen et~al.}]{bajaj2016ms}
Payal Bajaj, Daniel Campos, Nick Craswell, Li~Deng, Jianfeng Gao, Xiaodong Liu,
  Rangan Majumder, Andrew McNamara, Bhaskar Mitra, Tri Nguyen, et~al. 2016.
\newblock \href {https://arxiv.org/abs/1611.09268} {Ms marco: A human generated
  machine reading comprehension dataset}.
\newblock \emph{arXiv preprint arXiv:1611.09268}.

\bibitem[{Berant et~al.(2013)Berant, Chou, Frostig, and
  Liang}]{berant-etal-2013-semantic}
Jonathan Berant, Andrew Chou, Roy Frostig, and Percy Liang. 2013.
\newblock \href {https://aclanthology.org/D13-1160} {Semantic parsing on
  {F}reebase from question-answer pairs}.
\newblock In \emph{Proceedings of the 2013 Conference on Empirical Methods in
  Natural Language Processing}.

\bibitem[{Black et~al.(2021)Black, Leo, Wang, Leahy, and Biderman}]{gpt-neo}
Sid Black, Gao Leo, Phil Wang, Connor Leahy, and Stella Biderman. 2021.
\newblock \href {https://doi.org/10.5281/zenodo.5297715} {{GPT-Neo: Large Scale
  Autoregressive Language Modeling with Mesh-Tensorflow}}.

\bibitem[{Brown et~al.(2020)Brown, Mann, Ryder, Subbiah, Kaplan, Dhariwal,
  Neelakantan, Shyam, Sastry, Askell, Agarwal, Herbert-Voss, Krueger, Henighan,
  Child, Ramesh, Ziegler, Wu, Winter, Hesse, Chen, Sigler, Litwin, Gray, Chess,
  Clark, Berner, McCandlish, Radford, Sutskever, and
  Amodei}]{NEURIPS2020_1457c0d6}
Tom Brown, Benjamin Mann, Nick Ryder, Melanie Subbiah, Jared~D Kaplan, Prafulla
  Dhariwal, Arvind Neelakantan, Pranav Shyam, Girish Sastry, Amanda Askell,
  Sandhini Agarwal, Ariel Herbert-Voss, Gretchen Krueger, Tom Henighan, Rewon
  Child, Aditya Ramesh, Daniel Ziegler, Jeffrey Wu, Clemens Winter, Chris
  Hesse, Mark Chen, Eric Sigler, Mateusz Litwin, Scott Gray, Benjamin Chess,
  Jack Clark, Christopher Berner, Sam McCandlish, Alec Radford, Ilya Sutskever,
  and Dario Amodei. 2020.
\newblock \href
  {https://proceedings.neurips.cc/paper/2020/file/1457c0d6bfcb4967418bfb8ac142f64a-Paper.pdf}
  {Language models are few-shot learners}.
\newblock In \emph{Advances in Neural Information Processing Systems},
  volume~33.

\bibitem[{Chen et~al.(2017)Chen, Fisch, Weston, and Bordes}]{chen2017reading}
Danqi Chen, Adam Fisch, Jason Weston, and Antoine Bordes. 2017.
\newblock \href {https://www.aclweb.org/anthology/P17-1171} {Reading
  {Wikipedia} to answer open-domain questions}.
\newblock In \emph{Proceedings of the 55th Annual Meeting of the Association
  for Computational Linguistics (Volume 1: Long Papers)}.

\bibitem[{Chen et~al.(2021)Chen, Lakhotia, O{\u{g}}uz, Gupta, Lewis,
  Peshterliev, Mehdad, Gupta, and Yih}]{chen2021salient}
Xilun Chen, Kushal Lakhotia, Barlas O{\u{g}}uz, Anchit Gupta, Patrick Lewis,
  Stan Peshterliev, Yashar Mehdad, Sonal Gupta, and Wen-tau Yih. 2021.
\newblock \href {https://arxiv.org/abs/2110.06918} {Salient phrase aware dense
  retrieval: Can a dense retriever imitate a sparse one?}
\newblock \emph{arXiv preprint arXiv:2110.06918}.

\bibitem[{Chowdhery et~al.(2022)Chowdhery, Narang, Devlin, Bosma, Mishra,
  Roberts, Barham, Chung, Sutton, Gehrmann, Schuh, Shi, Tsvyashchenko, Maynez,
  Rao, Barnes, Tay, Shazeer, Prabhakaran, Reif, Du, Hutchinson, Pope, Bradbury,
  Austin, Isard, Gur-Ari, Yin, Duke, Levskaya, Ghemawat, Dev, Michalewski,
  Garc{\'i}a, Misra, Robinson, Fedus, Zhou, Ippolito, Luan, Lim, Zoph,
  Spiridonov, Sepassi, Dohan, Agrawal, Omernick, Dai, Pillai, Pellat,
  Lewkowycz, Moreira, Child, Polozov, Lee, Zhou, Wang, Saeta, Diaz, Firat,
  Catasta, Wei, Meier-Hellstern, Eck, Dean, Petrov, and
  Fiedel}]{Chowdhery2022PaLMSL}
Aakanksha Chowdhery, Sharan Narang, Jacob Devlin, Maarten Bosma, Gaurav Mishra,
  Adam Roberts, Paul Barham, Hyung~Won Chung, Charles Sutton, Sebastian
  Gehrmann, Parker Schuh, Kensen Shi, Sasha Tsvyashchenko, Joshua Maynez,
  Abhishek Rao, Parker Barnes, Yi~Tay, Noam~M. Shazeer, Vinodkumar Prabhakaran,
  Emily Reif, Nan Du, Benton~C. Hutchinson, Reiner Pope, James Bradbury, Jacob
  Austin, Michael Isard, Guy Gur-Ari, Pengcheng Yin, Toju Duke, Anselm
  Levskaya, Sanjay Ghemawat, Sunipa Dev, Henryk Michalewski, Xavier Garc{\'i}a,
  Vedant Misra, Kevin Robinson, Liam Fedus, Denny Zhou, Daphne Ippolito, David
  Luan, Hyeontaek Lim, Barret Zoph, Alexander Spiridonov, Ryan Sepassi, David
  Dohan, Shivani Agrawal, Mark Omernick, Andrew~M. Dai,
  Thanumalayan~Sankaranarayana Pillai, Marie Pellat, Aitor Lewkowycz,
  Erica~Oliveira Moreira, Rewon Child, Oleksandr Polozov, Katherine Lee,
  Zongwei Zhou, Xuezhi Wang, Brennan Saeta, Mark Diaz, Orhan Firat, Michele
  Catasta, Jason Wei, Kathleen~S. Meier-Hellstern, Douglas Eck, Jeff Dean, Slav
  Petrov, and Noah Fiedel. 2022.
\newblock \href {https://arxiv.org/abs/2204.02311} {{PaLM}: Scaling language
  modeling with pathways}.
\newblock \emph{ArXiv}, abs/2204.02311.

\bibitem[{de~Jong et~al.(2022)de~Jong, Zemlyanskiy, FitzGerald, Sha, and
  Cohen}]{jong2022mention}
Michiel de~Jong, Yury Zemlyanskiy, Nicholas FitzGerald, Fei Sha, and William~W.
  Cohen. 2022.
\newblock \href {https://openreview.net/forum?id=OY1A8ejQgEX} {Mention memory:
  incorporating textual knowledge into transformers through entity mention
  attention}.
\newblock In \emph{International Conference on Learning Representations}.

\bibitem[{Devlin et~al.(2019)Devlin, Chang, Lee, and
  Toutanova}]{devlin2019bert}
Jacob Devlin, Ming-Wei Chang, Kenton Lee, and Kristina Toutanova. 2019.
\newblock \href {https://www.aclweb.org/anthology/N19-1423} {{BERT}:
  Pre-training of deep bidirectional transformers for language understanding}.
\newblock In \emph{Proceedings of the 2019 Conference of the North {A}merican
  Chapter of the Association for Computational Linguistics: Human Language
  Technologies, Volume 1 (Long and Short Papers)}.

\bibitem[{Gao et~al.(2021)Gao, Yao, and Chen}]{gao-etal-2021-simcse}
Tianyu Gao, Xingcheng Yao, and Danqi Chen. 2021.
\newblock \href {https://doi.org/10.18653/v1/2021.emnlp-main.552} {{S}im{CSE}:
  Simple contrastive learning of sentence embeddings}.
\newblock In \emph{Proceedings of the 2021 Conference on Empirical Methods in
  Natural Language Processing}.

\bibitem[{Gururangan et~al.(2020)Gururangan, Marasovi{\'c}, Swayamdipta, Lo,
  Beltagy, Downey, and Smith}]{gururangan2020don}
Suchin Gururangan, Ana Marasovi{\'c}, Swabha Swayamdipta, Kyle Lo, Iz~Beltagy,
  Doug Downey, and Noah~A. Smith. 2020.
\newblock \href {https://doi.org/10.18653/v1/2020.acl-main.740} {Don{'}t stop
  pretraining: Adapt language models to domains and tasks}.
\newblock In \emph{Proceedings of the 58th Annual Meeting of the Association
  for Computational Linguistics}.

\bibitem[{Guu et~al.(2020)Guu, Lee, Tung, Pasupat, and Chang}]{guu2020realm}
Kelvin Guu, Kenton Lee, Zora Tung, Panupong Pasupat, and Mingwei Chang. 2020.
\newblock \href {http://proceedings.mlr.press/v119/guu20a.html} {Retrieval
  augmented language model pre-training}.
\newblock In \emph{Proceedings of the 37th International Conference on Machine
  Learning}.

\bibitem[{Izacard et~al.(2022)Izacard, Caron, Hosseini, Riedel, Bojanowski,
  Joulin, and Grave}]{izacard2021towards}
Gautier Izacard, Mathilde Caron, Lucas Hosseini, Sebastian Riedel, Piotr
  Bojanowski, Armand Joulin, and Edouard Grave. 2022.
\newblock \href {https://openreview.net/forum?id=jKN1pXi7b0} {Unsupervised
  dense information retrieval with contrastive learning}.
\newblock \emph{Transactions on Machine Learning Research}.

\bibitem[{Izacard and Grave(2021{\natexlab{a}})}]{izacard2021distilling}
Gautier Izacard and Edouard Grave. 2021{\natexlab{a}}.
\newblock \href {https://openreview.net/forum?id=NTEz-6wysdb} {Distilling
  knowledge from reader to retriever for question answering}.
\newblock In \emph{International Conference on Learning Representations}.

\bibitem[{Izacard and Grave(2021{\natexlab{b}})}]{izacard2021leveraging}
Gautier Izacard and Edouard Grave. 2021{\natexlab{b}}.
\newblock \href {https://www.aclweb.org/anthology/2021.eacl-main.74}
  {Leveraging passage retrieval with generative models for open domain question
  answering}.
\newblock In \emph{Proceedings of the 16th Conference of the European Chapter
  of the Association for Computational Linguistics: Main Volume}.

\bibitem[{Joshi et~al.(2017)Joshi, Choi, Weld, and
  Zettlemoyer}]{joshi2017triviaqa}
Mandar Joshi, Eunsol Choi, Daniel Weld, and Luke Zettlemoyer. 2017.
\newblock \href {https://www.aclweb.org/anthology/P17-1147} {{T}rivia{QA}: A
  large scale distantly supervised challenge dataset for reading
  comprehension}.
\newblock In \emph{Proceedings of the 55th Annual Meeting of the Association
  for Computational Linguistics (Volume 1: Long Papers)}.

\bibitem[{Ju et~al.(2021)Ju, Yang, and Wang}]{ju2021text}
Jia-Huei Ju, Jheng-Hong Yang, and Chuan-Ju Wang. 2021.
\newblock \href {https://arxiv.org/abs/2104.14133} {Text-to-text multi-view
  learning for passage re-ranking}.
\newblock In \emph{Proceedings of the 44th International ACM SIGIR Conference
  on Research and Development in Information Retrieval}.

\bibitem[{Karpukhin et~al.(2020)Karpukhin, O{\u{g}}uz, Min, Wu, Edunov, Chen,
  and Yih}]{karpukhin2020dense}
Vladimir Karpukhin, Barlas O{\u{g}}uz, Sewon Min, Ledell Wu, Sergey Edunov,
  Danqi Chen, and Wen-tau Yih. 2020.
\newblock \href {https://www.aclweb.org/anthology/2020.emnlp-main.550} {Dense
  passage retrieval for open-domain question answering}.
\newblock In \emph{Proceedings of the 2020 Conference on Empirical Methods in
  Natural Language Processing (EMNLP)}.

\bibitem[{Kingma and Ba(2015)}]{kingma2014adam}
Diederik~P Kingma and Jimmy Ba. 2015.
\newblock \href {https://arxiv.org/abs/1412.6980} {Adam: A method for
  stochastic optimization}.
\newblock In \emph{The 2015 International Conference for Learning
  Representations}.

\bibitem[{Kwiatkowski et~al.(2019)Kwiatkowski, Palomaki, Redfield, Collins,
  Parikh, Alberti, Epstein, Polosukhin, Kelcey, Devlin, Lee, Toutanova, Jones,
  Chang, Dai, Uszkoreit, Le, and Petrov}]{Kwiatkowski2019natural}
Tom Kwiatkowski, Jennimaria Palomaki, Olivia Redfield, Michael Collins, Ankur
  Parikh, Chris Alberti, Danielle Epstein, Illia Polosukhin, Matthew Kelcey,
  Jacob Devlin, Kenton Lee, Kristina~N. Toutanova, Llion Jones, Ming-Wei Chang,
  Andrew Dai, Jakob Uszkoreit, Quoc Le, and Slav Petrov. 2019.
\newblock \href {https://www.aclweb.org/anthology/Q19-1026/} {Natural
  questions: a benchmark for question answering research}.
\newblock \emph{Transactions of the Association of Computational Linguistics}.

\bibitem[{Lee et~al.(2019)Lee, Chang, and Toutanova}]{lee-etal-2019-latent}
Kenton Lee, Ming-Wei Chang, and Kristina Toutanova. 2019.
\newblock \href {https://www.aclweb.org/anthology/P19-1612} {Latent retrieval
  for weakly supervised open domain question answering}.
\newblock In \emph{Proceedings of the 57th Annual Meeting of the Association
  for Computational Linguistics}.

\bibitem[{Lester et~al.(2021)Lester, Al-Rfou, and
  Constant}]{lester-etal-2021-power}
Brian Lester, Rami Al-Rfou, and Noah Constant. 2021.
\newblock \href {https://doi.org/10.18653/v1/2021.emnlp-main.243} {The power of
  scale for parameter-efficient prompt tuning}.
\newblock In \emph{Proceedings of the 2021 Conference on Empirical Methods in
  Natural Language Processing}.

\bibitem[{Lewis et~al.(2020)Lewis, Perez, Piktus, Petroni, Karpukhin, Goyal,
  K\"{u}ttler, Lewis, Yih, Rockt\"{a}schel, Riedel, and Kiela}]{lewis20rag}
Patrick Lewis, Ethan Perez, Aleksandra Piktus, Fabio Petroni, Vladimir
  Karpukhin, Naman Goyal, Heinrich K\"{u}ttler, Mike Lewis, Wen-tau Yih, Tim
  Rockt\"{a}schel, Sebastian Riedel, and Douwe Kiela. 2020.
\newblock \href
  {https://proceedings.neurips.cc/paper/2020/file/6b493230205f780e1bc26945df7481e5-Paper.pdf}
  {Retrieval-augmented generation for knowledge-intensive nlp tasks}.
\newblock In \emph{Advances in Neural Information Processing Systems}.

\bibitem[{Lin et~al.(2021)Lin, Ma, Lin, Yang, Pradeep, and
  Nogueira}]{Lin_etal_SIGIR2021_Pyserini}
Jimmy Lin, Xueguang Ma, Sheng-Chieh Lin, Jheng-Hong Yang, Ronak Pradeep, and
  Rodrigo Nogueira. 2021.
\newblock \href {https://dl.acm.org/doi/10.1145/3404835.3463238} {{Pyserini}: A
  {Python} toolkit for reproducible information retrieval research with sparse
  and dense representations}.
\newblock In \emph{Proceedings of the 44th Annual International ACM SIGIR
  Conference on Research and Development in Information Retrieval (SIGIR
  2021)}.

\bibitem[{Ma et~al.(2021)Ma, Sun, Pradeep, and Lin}]{ma2021replication}
Xueguang Ma, Kai Sun, Ronak Pradeep, and Jimmy Lin. 2021.
\newblock \href {https://arxiv.org/abs/2104.05740} {A replication study of
  dense passage retriever}.
\newblock \emph{arXiv preprint arXiv:2104.05740}.

\bibitem[{Min et~al.(2022)Min, Lewis, Zettlemoyer, and
  Hajishirzi}]{min2021metaicl}
Sewon Min, Mike Lewis, Luke Zettlemoyer, and Hannaneh Hajishirzi. 2022.
\newblock \href {https://doi.org/10.18653/v1/2022.naacl-main.201} {{M}eta{ICL}:
  Learning to learn in context}.
\newblock In \emph{Proceedings of the 2022 Conference of the North American
  Chapter of the Association for Computational Linguistics: Human Language
  Technologies}.

\bibitem[{Nogueira et~al.(2020)Nogueira, Jiang, Pradeep, and
  Lin}]{nogueira-etal-2020-document}
Rodrigo Nogueira, Zhiying Jiang, Ronak Pradeep, and Jimmy Lin. 2020.
\newblock \href {https://doi.org/10.18653/v1/2020.findings-emnlp.63} {Document
  ranking with a pretrained sequence-to-sequence model}.
\newblock In \emph{Findings of the Association for Computational Linguistics:
  EMNLP 2020}.

\bibitem[{Nogueira~dos Santos et~al.(2020)Nogueira~dos Santos, Ma, Nallapati,
  Huang, and Xiang}]{nogueira-dos-santos-etal-2020-beyond}
Cicero Nogueira~dos Santos, Xiaofei Ma, Ramesh Nallapati, Zhiheng Huang, and
  Bing Xiang. 2020.
\newblock \href {https://doi.org/10.18653/v1/2020.emnlp-main.134} {Beyond
  [{CLS}] through ranking by generation}.
\newblock In \emph{Proceedings of the 2020 Conference on Empirical Methods in
  Natural Language Processing (EMNLP)}.

\bibitem[{Paszke et~al.(2019)Paszke, Gross, Massa, Lerer, Bradbury, Chanan,
  Killeen, Lin, Gimelshein, Antiga, Desmaison, Kopf, Yang, DeVito, Raison,
  Tejani, Chilamkurthy, Steiner, Fang, Bai, and Chintala}]{paszke2019pytorch}
Adam Paszke, Sam Gross, Francisco Massa, Adam Lerer, James Bradbury, Gregory
  Chanan, Trevor Killeen, Zeming Lin, Natalia Gimelshein, Luca Antiga, Alban
  Desmaison, Andreas Kopf, Edward Yang, Zachary DeVito, Martin Raison, Alykhan
  Tejani, Sasank Chilamkurthy, Benoit Steiner, Lu~Fang, Junjie Bai, and Soumith
  Chintala. 2019.
\newblock \href
  {https://proceedings.neurips.cc/paper/2019/file/bdbca288fee7f92f2bfa9f7012727740-Paper.pdf}
  {Pytorch: An imperative style, high-performance deep learning library}.
\newblock In \emph{Advances in Neural Information Processing Systems}.

\bibitem[{Ponte and Croft(1998)}]{10.1145/290941.291008}
Jay~M. Ponte and W.~Bruce Croft. 1998.
\newblock \href {https://doi.org/10.1145/290941.291008} {A language modeling
  approach to information retrieval}.
\newblock In \emph{Proceedings of the 21st Annual International ACM SIGIR
  Conference on Research and Development in Information Retrieval}.

\bibitem[{Radford et~al.(2019)Radford, Wu, Child, Luan, Amodei, and
  Sutskever}]{radford2019language}
Alec Radford, Jeff Wu, Rewon Child, David Luan, Dario Amodei, and Ilya
  Sutskever. 2019.
\newblock Language models are unsupervised multitask learners.

\bibitem[{Rae et~al.(2021)Rae, Borgeaud, Cai, Millican, Hoffmann, Song,
  Aslanides, Henderson, Ring, Young, Rutherford, Hennigan, Menick, Cassirer,
  Powell, van~den Driessche, Hendricks, Rauh, Huang, Glaese, Welbl, Dathathri,
  Huang, Uesato, Mellor, Higgins, Creswell, McAleese, Wu, Elsen, Jayakumar,
  Buchatskaya, Budden, Sutherland, Simonyan, Paganini, Sifre, Martens, Li,
  Kuncoro, Nematzadeh, Gribovskaya, Donato, Lazaridou, Mensch, Lespiau,
  Tsimpoukelli, Grigorev, Fritz, Sottiaux, Pajarskas, Pohlen, Gong, Toyama,
  de~Masson~d'Autume, Li, Terzi, Mikulik, Babuschkin, Clark, de~Las~Casas, Guy,
  Jones, Bradbury, Johnson, Hechtman, Weidinger, Gabriel, Isaac, Lockhart,
  Osindero, Rimell, Dyer, Vinyals, Ayoub, Stanway, Bennett, Hassabis,
  Kavukcuoglu, and Irving}]{Rae2021ScalingLM}
Jack~W. Rae, Sebastian Borgeaud, Trevor Cai, Katie Millican, Jordan Hoffmann,
  Francis Song, John Aslanides, Sarah Henderson, Roman Ring, Susannah Young,
  Eliza Rutherford, Tom Hennigan, Jacob Menick, Albin Cassirer, Richard Powell,
  George van~den Driessche, Lisa~Anne Hendricks, Maribeth Rauh, Po-Sen Huang,
  Amelia Glaese, Johannes Welbl, Sumanth Dathathri, Saffron Huang, Jonathan
  Uesato, John F.~J. Mellor, Irina Higgins, Antonia Creswell, Nathan McAleese,
  Amy Wu, Erich Elsen, Siddhant~M. Jayakumar, Elena Buchatskaya, David Budden,
  Esme Sutherland, Karen Simonyan, Michela Paganini, L.~Sifre, Lena Martens,
  Xiang~Lorraine Li, Adhiguna Kuncoro, Aida Nematzadeh, Elena Gribovskaya,
  Domenic Donato, Angeliki Lazaridou, Arthur Mensch, Jean-Baptiste Lespiau,
  Maria Tsimpoukelli, N.~K. Grigorev, Doug Fritz, Thibault Sottiaux, Mantas
  Pajarskas, Tobias Pohlen, Zhitao Gong, Daniel Toyama, Cyprien
  de~Masson~d'Autume, Yujia Li, Tayfun Terzi, Vladimir Mikulik, Igor
  Babuschkin, Aidan Clark, Diego de~Las~Casas, Aurelia Guy, Chris Jones, James
  Bradbury, Matthew~G. Johnson, Blake~A. Hechtman, Laura Weidinger, Iason
  Gabriel, William~S. Isaac, Edward Lockhart, Simon Osindero, Laura Rimell,
  Chris Dyer, Oriol Vinyals, Kareem~W. Ayoub, Jeff Stanway, L.~L. Bennett,
  Demis Hassabis, Koray Kavukcuoglu, and Geoffrey Irving. 2021.
\newblock \href {https://arxiv.org/abs/2112.11446} {Scaling language models:
  Methods, analysis \& insights from training gopher}.
\newblock \emph{ArXiv}, abs/2112.11446.

\bibitem[{Raffel et~al.(2020)Raffel, Shazeer, Roberts, Lee, Narang, Matena,
  Zhou, Li, and Liu}]{raffel2020t5}
Colin Raffel, Noam Shazeer, Adam Roberts, Katherine Lee, Sharan Narang, Michael
  Matena, Yanqi Zhou, Wei Li, and Peter~J. Liu. 2020.
\newblock \href {http://jmlr.org/papers/v21/20-074.html} {Exploring the limits
  of transfer learning with a unified text-to-text transformer}.
\newblock \emph{Journal of Machine Learning Research}, 21(140):1--67.

\bibitem[{Rajpurkar et~al.(2016)Rajpurkar, Zhang, Lopyrev, and
  Liang}]{rajpurkar-etal-2016-squad}
Pranav Rajpurkar, Jian Zhang, Konstantin Lopyrev, and Percy Liang. 2016.
\newblock \href {https://doi.org/10.18653/v1/D16-1264} {{SQ}u{AD}: 100,000+
  questions for machine comprehension of text}.
\newblock In \emph{Proceedings of the 2016 Conference on Empirical Methods in
  Natural Language Processing}.

\bibitem[{Robertson and Zaragoza(2009)}]{Robertson2009bm25}
Stephen Robertson and Hugo Zaragoza. 2009.
\newblock \href {https://doi.org/10.1561/1500000019} {The probabilistic
  relevance framework: Bm25 and beyond}.
\newblock \emph{Foundations and Trends in Information Retrieval}.

\bibitem[{Sachan et~al.(2021{\natexlab{a}})Sachan, Patwary, Shoeybi, Kant,
  Ping, Hamilton, and Catanzaro}]{sachan2021end}
Devendra~Singh Sachan, Mostofa Patwary, Mohammad Shoeybi, Neel Kant, Wei Ping,
  William~L Hamilton, and Bryan Catanzaro. 2021{\natexlab{a}}.
\newblock \href {https://aclanthology.org/2021.acl-long.519/} {End-to-end
  training of neural retrievers for open-domain question answering}.
\newblock In \emph{Joint Conference of the 59th Annual Meeting of the
  Association for Computational Linguistics and the 11th International Joint
  Conference on Natural Language Processing (ACL-IJCNLP)}.

\bibitem[{Sachan et~al.(2021{\natexlab{b}})Sachan, Reddy, Hamilton, Dyer, and
  Yogatama}]{sachan2021endtoend}
Devendra~Singh Sachan, Siva Reddy, William~L. Hamilton, Chris Dyer, and Dani
  Yogatama. 2021{\natexlab{b}}.
\newblock \href {https://openreview.net/forum?id=5KWmB6JePx} {End-to-end
  training of multi-document reader and retriever for open-domain question
  answering}.
\newblock In \emph{Advances in Neural Information Processing Systems}.

\bibitem[{Sanh et~al.(2022)Sanh, Webson, Raffel, Bach, Sutawika, Alyafeai,
  Chaffin, Stiegler, Raja, Dey, Bari, Xu, Thakker, Sharma, Szczechla, Kim,
  Chhablani, Nayak, Datta, Chang, Jiang, Wang, Manica, Shen, Yong, Pandey,
  Bawden, Wang, Neeraj, Rozen, Sharma, Santilli, Fevry, Fries, Teehan, Scao,
  Biderman, Gao, Wolf, and Rush}]{sanh2022multitask}
Victor Sanh, Albert Webson, Colin Raffel, Stephen Bach, Lintang Sutawika, Zaid
  Alyafeai, Antoine Chaffin, Arnaud Stiegler, Arun Raja, Manan Dey, M~Saiful
  Bari, Canwen Xu, Urmish Thakker, Shanya~Sharma Sharma, Eliza Szczechla,
  Taewoon Kim, Gunjan Chhablani, Nihal Nayak, Debajyoti Datta, Jonathan Chang,
  Mike Tian-Jian Jiang, Han Wang, Matteo Manica, Sheng Shen, Zheng~Xin Yong,
  Harshit Pandey, Rachel Bawden, Thomas Wang, Trishala Neeraj, Jos Rozen,
  Abheesht Sharma, Andrea Santilli, Thibault Fevry, Jason~Alan Fries, Ryan
  Teehan, Teven~Le Scao, Stella Biderman, Leo Gao, Thomas Wolf, and Alexander~M
  Rush. 2022.
\newblock \href {https://openreview.net/forum?id=9Vrb9D0WI4} {Multitask
  prompted training enables zero-shot task generalization}.
\newblock In \emph{International Conference on Learning Representations}.

\bibitem[{Sciavolino et~al.(2021)Sciavolino, Zhong, Lee, and
  Chen}]{sciavolino2021simple}
Christopher Sciavolino, Zexuan Zhong, Jinhyuk Lee, and Danqi Chen. 2021.
\newblock \href {https://arxiv.org/abs/2109.08535} {Simple entity-centric
  questions challenge dense retrievers}.
\newblock In \emph{Empirical Methods in Natural Language Processing (EMNLP)}.

\bibitem[{Smith et~al.(2022)Smith, Patwary, Norick, LeGresley, Rajbhandari,
  Casper, Liu, Prabhumoye, Zerveas, Korthikanti et~al.}]{smith2022using}
Shaden Smith, Mostofa Patwary, Brandon Norick, Patrick LeGresley, Samyam
  Rajbhandari, Jared Casper, Zhun Liu, Shrimai Prabhumoye, George Zerveas,
  Vijay Korthikanti, et~al. 2022.
\newblock \href {https://arxiv.org/abs/2201.11990} {Using deepspeed and
  megatron to train megatron-turing nlg 530b, a large-scale generative language
  model}.
\newblock \emph{arXiv preprint arXiv:2201.11990}.

\bibitem[{Thakur et~al.(2021)Thakur, Reimers, R{\"u}ckl{\'e}, Srivastava, and
  Gurevych}]{thakur2021beir}
Nandan Thakur, Nils Reimers, Andreas R{\"u}ckl{\'e}, Abhishek Srivastava, and
  Iryna Gurevych. 2021.
\newblock \href {https://openreview.net/forum?id=wCu6T5xFjeJ} {{BEIR}: A
  heterogeneous benchmark for zero-shot evaluation of information retrieval
  models}.
\newblock In \emph{Thirty-fifth Conference on Neural Information Processing
  Systems Datasets and Benchmarks Track (Round 2)}.

\bibitem[{Wei et~al.(2022)Wei, Bosma, Zhao, Guu, Yu, Lester, Du, Dai, and
  Le}]{wei2022finetuned}
Jason Wei, Maarten Bosma, Vincent Zhao, Kelvin Guu, Adams~Wei Yu, Brian Lester,
  Nan Du, Andrew~M. Dai, and Quoc~V Le. 2022.
\newblock \href {https://openreview.net/forum?id=gEZrGCozdqR} {Finetuned
  language models are zero-shot learners}.
\newblock In \emph{International Conference on Learning Representations}.

\bibitem[{Wolf et~al.(2020)Wolf, Debut, Sanh, Chaumond, Delangue, Moi, Cistac,
  Rault, Louf, Funtowicz, Davison, Shleifer, von Platen, Ma, Jernite, Plu, Xu,
  Le~Scao, Gugger, Drame, Lhoest, and Rush}]{wolf-etal-2020-transformers}
Thomas Wolf, Lysandre Debut, Victor Sanh, Julien Chaumond, Clement Delangue,
  Anthony Moi, Pierric Cistac, Tim Rault, Remi Louf, Morgan Funtowicz, Joe
  Davison, Sam Shleifer, Patrick von Platen, Clara Ma, Yacine Jernite, Julien
  Plu, Canwen Xu, Teven Le~Scao, Sylvain Gugger, Mariama Drame, Quentin Lhoest,
  and Alexander Rush. 2020.
\newblock \href {https://doi.org/10.18653/v1/2020.emnlp-demos.6} {Transformers:
  State-of-the-art natural language processing}.
\newblock In \emph{Proceedings of the 2020 Conference on Empirical Methods in
  Natural Language Processing: System Demonstrations}.

\bibitem[{Zhai and Lafferty(2001)}]{10.1145/383952.384019}
Chengxiang Zhai and John Lafferty. 2001.
\newblock \href {https://doi.org/10.1145/383952.384019} {A study of smoothing
  methods for language models applied to ad hoc information retrieval}.
\newblock In \emph{Proceedings of the 24th Annual International ACM SIGIR
  Conference on Research and Development in Information Retrieval}.

\bibitem[{Zhou and Devlin(2021)}]{zhou-devlin-2021-multi}
Giulio Zhou and Jacob Devlin. 2021.
\newblock \href {https://doi.org/10.18653/v1/2021.emnlp-main.443} {Multi-vector
  attention models for deep re-ranking}.
\newblock In \emph{Proceedings of the 2021 Conference on Empirical Methods in
  Natural Language Processing}.

\end{thebibliography}
\bibliographystyle{acl_natbib}

\clearpage
\appendix


\section{Appendix}
\label{sec:appendix}

\subsection{Training Hyperparameters}      \label{appendix:train-hparams}

\paragraph{Supervised Retriever}
We use Adam optimizer~\cite{kingma2014adam}, a batch size of 128, 1 hard negative example for each positive pair, a learning rate of 2e-5 with a linear decay, weight decay of 0.1, and train for 3 epochs on SQuAD-Open, 40 epochs for NQ and TriviaQA, and 20 epochs on WebQ. Model training was performed on 16 GPUs.

\paragraph{Fusion-in-Decoder Reader}
We use Adam optimizer~\cite{kingma2014adam}, a batch size of 64, a learning rate of 2e-5 with a linear decay, a weight decay of 0.1, gradient clipping with a maximum value of 1.0, and train for 3 epochs on SQuAD-Open, 10 epochs for NQ and TriviaQA. Model training was performed on 64 GPUs. For our experiments, we use the Fusion-in-Decoder model implementation from the open-source repository (\url{https://github.com/DevSinghSachan/emdr2})~\cite{sachan2021endtoend}.

\subsection{Instruction Prompt Selection} \label{appendix:instruction-prompt}

\begin{table*}[!t]
\small
\centering
\begin{tabular}{@{}l | l c c c c @{}}
 \toprule
 \tf{Retriever /} & \tf{Instruction Prompt} & \multicolumn{4}{c}{\tf{NQ (dev)}}  \\
 \tf{Re-Ranker}   &                &  Top-1 & Top-5 & Top-20 & Top-100  \\
\midrule
       BM25       &  & 22.3  & 43.8  & 62.3   &  76.0  \\  
\cmidrule{2-6}
\phantom{BM25} + UPR &    `'      & 28.3 & 56.1 & 73.2 & 82.4 \\
  \phantom{BM25} + UPR & `\textit{Score the following question based on this passage.}' & 35.3 & 62.6 & 76.4 & 83.0 \\
  \phantom{BM25} + UPR & `\textit{A possible question based on this passage is.}' & 33.8 & 61.6 & 76.2 & \tf{83.1} \\
  \phantom{BM25} + UPR & `\textit{This is a relevant document for the following question.}' & 33.7 & 61.8 & 76.0 & 83.0 \\
  \phantom{BM25} + UPR & `\textit{Please write a question based on this passage.}' & \tf{36.1} & \tf{62.8} & \tf{76.8} & \tf{83.1} \\
\bottomrule
\end{tabular}
\caption{
Comparison of different instruction prompts when applied to UPR framework and evaluated on the NQ development set. Results highlight that UPR works better with simple instructions. Best results are highlighted in bold.
}
\label{tab:qa-prompts}
\end{table*}

We cross-validate using several prompts formulated as natural language instructions to aid in question reconstruction.
We re-rank top-1000 BM25 passages of NQ development set using different instructions including the case with no instruction.
Results in Table~\ref{tab:qa-prompts} reveal that when prompted via instructions, PLMs perform better than the case when not given any instructions. We also note that simple but effective instructions can lead to a higher top-1 accuracy. Due to its better accuracy, we use the instruction "\textit{Please write a question based on this passage}" for all the experiments in this paper.

\subsection{Analysis}
\label{subsec:analysis}

\begin{table*}[t]
\begin{minipage}{\linewidth}
\small
\centering
\begin{tabular}{p{3cm} p{6cm} p{6cm}}
\toprule
\textbf{Question / Answer}  & \textbf{BM25} & \textbf{BM25 + UPR} \\
\midrule
\makecell[t{p{3cm}}]{\textbf{Question}: when did beauty and the beast come out on dvd? \\ \textbf{Answer}: June 6 , 2017} & 
\makecell[t{p{6cm}}]{{\textbf{Title}}: Beauty and the Beast: The Enchanted Christmas \\ \textbf{Passage}: these traits were incorporated into Forte, the pipe organ, who did not want the Beast to become human again. Unlike the other characters, Forte was animated entirely by computers. The film was first released on VHS on November 11, 1997. A bare-bones DVD was released on October 13, 1998. Both editions were quickly taken out of print and the film remained unavailable until Disney released the Special Edition DVD and VHS on November 12, 2002, just after the studio released the original film\'s Special Edition DVD release. The new DVD featured a remake music video of the song "As Long'} & 
\makecell[t{p{6cm}}]{\textbf{Title}: Beauty and the Beast (2017 film) \\ \textbf{Passage}: Golden Globe Awards. Disney released the final trailer on January 30, 2017. "Beauty and the Beast" was released on Blu-ray, DVD and Digital HD on \textbf{June 6, 2017}. The film debuted at No. 1 on the NPD VideoScan overall disc sales chart, with all other titles in the top 20, collectively, selling only 40\% as many units as "Beauty and the Beast". The movie regained the top spot on the national home video sales charts during its third week of release. The movie became available on Netflix on September 19, 2017. "Beauty and the Beast" grossed \$504 million in the} \\
\midrule
\makecell[t{p{3cm}}]{\textbf{Question}: who sings i 'm back in the saddle again? \\ \textbf{Answer}: Gene Autry} & \makecell[t{p{6cm}}]{\textbf{Title}: Back in the Saddle (film) \\ \textbf{Passage}: remember her as "an interesting actress" who had talent, but "seemed doomed to work in nothing but B-westerns." Remembering her work on "Back in the Saddle", Wells would later recall, "Actually, I didn\'t have much to do with Gene in the film, I had more scenes with Edward Norris." Regarding the singing sequences, she remembered, "Usually I was dubbed but, occasionally, if it wasn\'t something too difficult, I was allowed to do it. They prerecord the songs, then a huge machine comes onto the stage and you lip-sync to the recording." "Back in the Saddle" was filmed January 21 to}
& 
\makecell[t{p{6cm}}]{\textbf{Title}: Back in the Saddle Again \\ \textbf{Passage}: "Gene Autry\'s Melody Ranch" which premiered over the CBS Radio Network on January 7, 1940 where the show ran until 1956. The song also became the title song for the Autry film "Back in the Saddle" (Republic Pictures, March 14, 1941). \textbf{Gene Autry} recorded "Back in the Saddle Again" for the first time on April 18, 1939 in Los Angeles for Columbia Record Corporation, matrix number LA 1865, which was originally issued on Vocalion 05080. LA 1865 also issued on the Conqueror, OKeh, and Columbia labels. Early Vocalion and Conqueror labels say "BACK TO THE SADDLE". Conqueror was a private.} \\
\midrule
\makecell[t{p{3cm}}]{\textbf{Question}: who did the astros beat in the world series? \\ \textbf{Answer}: Los Angeles Dodgers} & \makecell[t{p{6cm}}]{\textbf{Title}: History of the Houston Astros. \\ \textbf{Passage}: starters Ezequiel Astacio and Wandy Rodríguez were also successful. In July alone, the Astros went 22–7, the best single month record in the club's history. The Astros finished the 2005 regular season by winning a wild card berth on the final day of the regular season, just as they did in 2004, becoming only the second team to come from 15 games under .500 to enter the post season, the other team being the 1914 Boston Braves, now the Atlanta Braves. (Those Braves would go on and sweep the Philadelphia Athletics in the World Series. Coincidentally, the Astros beat out} & 
\makecell[t{p{6cm}}]{\textbf{Title}: 2017 Houston Astros season \\ \textbf{Passage}: in four games. Houston then advanced to the AL Championship Series (ALCS) and defeated the New York Yankees in seven games for their first American League pennant. It was the second league championship in franchise history, and first since 2005 and they became the first team in history to make it to the World Series as members of both the National League and the American League. Finally, the Astros faced and defeated the \textbf{Los Angeles Dodgers} in seven games in the World Series, garnering the first World Series title in franchise history. During the regular season, the Astros featured the} \\ 
\midrule
\makecell[t{p{3cm}}]{\textbf{Question}: who won the big 10 football championship in 2016? \\ \textbf{Answer}: Penn State Nittany Lions} & \makecell[t{p{6cm}}]{\textbf{Title}: 2016 Big Ten Football Championship Game \\ \textbf{Passage}: 2016 Big Ten Football Championship Game The 2016 Big Ten Football Championship Game was played December 3, 2016 at Lucas Oil Stadium in Indianapolis, Indiana. It was the sixth annual Big Ten Football Championship Game to determine the 2016 champion of the Big Ten Conference. The 2016 Big Ten Championship Game pitted the Wisconsin Badgers, champions of the West Division, who made its fourth appearance in six years in the conference title game, against the East Division champion \textbf{Penn State Nittany Lions}, who made their first-ever appearance in the conference championship game. Penn State and Ohio State had identical 8–1} & \makecell[t{p{6cm}}]{\textbf{Title}: 2016 Big Ten Conference football season \\ \textbf{Passage}: since the conference instituted divisions. Wisconsin won the West Division for the fourth time in the six years the division had existed. In the Big Ten Championship held on December 3, 2016 at Lucas Oil Stadium in Indianapolis, Indiana, \textit{\textbf{Penn State}} defeated Wisconsin 38–31 to win the Big Ten. Several Big Ten teams changed head coaches in 2016. Tracy Claeys at Minnesota had the "interim" tag removed from his title and served as the permanent head coach. D. J. Durkin was the new head coach at Maryland taking over for Randy Edsall after having spent the previous year as the} \\

\bottomrule
\end{tabular}
\caption{Selected examples from the NQ development set of the top-1 retrieved passage from BM25 and the top passage obtained by UPR re-ranking of 1000 passages. If the answer exists in the passage it is highlighted in bold.
UPR leverages powerful cross-attention between the question and passage tokens and hence is able to obtain improved passage rankings.
}
\label{table:analysis_examples}
\end{minipage}
\end{table*}
In Table~\ref{table:analysis_examples}, we present some examples of questions and their BM25 retrieved and UPR re-ranked top-1 passages.
While BM25 retrieves passages with high lexical overlap, UPR owing to its cross-attention mechanism is more able to understand the relationships between tokens in the question and passage and thus leads to an improvement in passage rankings over the first-stage retriever.
In the last example, we note that although the BM25 retrieved passage contains the ground-truth answer, it should be considered a false positive result. On the other hand, UPR leads to the correctly ranked passage but the exact match evaluation metric marks it as incorrect as it does not match the full ground-truth answer.

\subsection{BEIR Benchmark Results}        \label{appendix:beir-benchmark}


\begin{table*}[t]
\addtolength{\tabcolsep}{-0.45pt}
\small
\centering
\begin{tabular}{@{}l c c |c c c c| c c c c@{}}
 \toprule
 \multirow{3}{*}{\tf{Dataset}} & \multirow{3}{*}{\tf{\#Q}} & \multirow{3}{*}{\tf{\#E}} & \multicolumn{4}{c|}{\tf{NDCG@10}} & \multicolumn{4}{c}{\tf{Recall@100}} \\
 \cmidrule{4-11}
  & & & \multicolumn{2}{c}{\textbf{BM25}} & \multicolumn{2}{c|}{\textbf{Contriever}} & \multicolumn{2}{c}{\textbf{BM25}} & \multicolumn{2}{c}{\textbf{Contriever}} \\
                  &            &             & original & re-ranked & original & re-ranked & original & re-ranked & original & re-ranked  \\
\midrule
Scifact & \phantom{00}300        & \phantom{0}5K & 66.5 & 70.3 & 64.9 & 69.6 & 90.8 & 94.2 & 92.6 & 94.3 \\
Scidocs & \phantom{0}1000        & 25K           & 15.8 & 17.0 & 14.9 & 17.3 & 35.6 & 39.0 & 36.0 & 39.0 \\
Nfcorpus & \phantom{00}323       & 3.5K          & 32.5 &  34.8   & 31.7 & 33.3 & 25.0 &  28.0   & 29.0 & 31.3 \\
FIQA-2018 & \phantom{00}648      & 57K           & 23.6 & 44.4 & 24.5 & 45.0 & 53.9 & 67.7 & 56.2 & 72.8 \\
Trec-covid & \phantom{000}50     & 0.2M          & 65.5 & 68.8 & 27.4 & 60.4 & 49.8 & 54.8 & 17.2 & 36.7 \\
Touche-2020 & \phantom{000}49    & 0.4M          & 36.8 & \red{20.6} & 19.3 & 21.3 & 53.8 & \red{45.7} & 22.5 & 42.4 \\
NQ & \phantom{0}3452             & 2.7M          & 32.9 & 45.4 & 25.4 & 44.2 & 76.0 & 87.7 & 77.1 & 88.4 \\
MS-Marco & \phantom{0}6980       & 8.8M          & 22.8 & 30.2 & 20.6 & 30.7 & 65.8 & 76.9 & 67.2 & 79.1 \\
HotpotQA & \phantom{0}7405       & 5.2M          & 60.3 & 73.3 & 48.1 & 72.2 & 74.0 & 82.5 & 70.4 & 80.8 \\
ArguAna & \phantom{0}1406        & 8.7K          & 31.5 & 37.2 & 37.9 & 50.3 & 94.2 & 98.2 & 90.1 & 97.5 \\
CQADupStack & 13145              & 0.5M          & 29.9 & 41.6 & 28.4 & 41.7 & 60.6 & 70.1 & 61.4 & 71.3 \\
Quora & 10000                    & 0.5M          & 78.9 & 83.1 & 83.5 & \red{82.8} & 97.3 & 98.8 & 98.7 & 98.9 \\
DBpedia & \phantom{00}400        & 4.6M          & 31.3 & 35.4 & 29.2 & 33.8 & 39.8 & 53.3 & 45.3 & 47.8 \\
Fever & \phantom{0}6666          & 5.4M          & 75.3 & \red{59.1} & 68.2 & \red{57.3} & 93.1 & \red{84.2} & 93.6 & \red{83.1} \\
Climate-Fever & \phantom{0}1535  & 5.4M          & 21.3 & \red{11.7} & 15.5 & \phantom{0}\red{9.5} & 43.6 & \red{39.2} & 44.1 & \red{31.3} \\
\midrule
Average       &                  &  & 41.6 & \tf{44.9} & 36.0 & 44.6 & 63.6 & \tf{68.0} & 60.1 & 66.3 \\
\bottomrule
\end{tabular}
\caption{UPR re-ranking results on the BEIR benchmark~\cite{thakur2021beir}.
\#Q and \#E denotes the number of queries and evidence documents, respectively.
Upon re-ranking the top-1000 documents with the T0-3B language model, on average, the performance of both BM25 and Contriever improve on the NDCG@10 and Recall@100 metrics.
We also observe a drop in scores on some datasets which is highlighted in red.
}
\label{tab:beir-benchmark}
\end{table*}

We re-rank the top-1000 documents from the BM25 and Contriever retrievers with the T0-3B pre-trained language model and evaluate performance with NDCG@10 and Recall@100 metrics.
We present the results of the individual datasets included in the BEIR benchmark in Table~\ref{tab:beir-benchmark}.
On both the metrics, the initial scores of BM25 are much higher than those of Contriever.
After re-ranking, BM25 retriever obtains improvements on 12 out of 15 datasets while Contriever obtains improvements on 13 out of 15 datasets. 
On average, NDCG@10 scores improve by 3-8\% and Recall@100 improves by 5-6\%. 
The performance gap between BM25 and Contriever also narrows down after re-ranking.

Due to the diversity in datasets, there is a considerable variation in performance gains across them. 
In the case of BM25, the highest relative performance gains are obtained by UPR on datasets containing information-seeking questions such as FIQA-2018, NQ, MS-Marco, etc. Similarly, for Contriever, the relative gains are much higher for the datasets of Trec-Covid, NQ, HotpotQA, etc., where the queries are questions. On other datasets, the relative gains from re-ranking are moderate to little.

For both the retrievers, we also observe a drop in performance on the fact-verification datasets of Fever and Climate-fever (results highlighted in red color in Table~\ref{tab:beir-benchmark}). In addition, re-ranking BM25 also results in a drop in performance on the Touche-2020 dataset.
We note that in these datasets, the queries are statements such as claims, which presents a different retrieval challenge for re-ranking. 
We anticipate that by experimenting with different prompt instructions in UPR to better suit the end-task and cross-validating with the number of top-K documents to be re-ranked, results can be improved on these datasets.
However, we leave these explorations as a part of future work.


\section{Reproducibility Checklist}

\subsection{For all reported experimental results}
\begin{itemize}
	\item \textit{A clear description of the mathematical setting, algorithm, and/or model}: This is provided in the main paper in Sec.~\ref{sec:method}.

	\item \textit{A link to a downloadable source code, with specification of all dependencies, including external libraries}: We are submitting the source codes as a zip file.

	\item \textit{A description of computing infrastructure used}: We run experiments on a cluster containing V100 GPUs where each node's specifications are: Number of CPUs: 256, Physical Memory: 1.2TB, GPU model: 8 x Nvidia V100, GPU architecture and memory: Volta/32GB, Arch: x86\_64, and Disk size: 4TB. For experiments in Sec.~\ref{subsec:latency}, we used a single node of 8 x A100 GPUs of 40GB memory.

	\item \textit{The average runtime for each model or algorithm (e.g., training, inference, etc.), or estimated energy cost}: We discuss the average runtime of performing inference with UPR in Sec.~\ref{subsec:latency}. However, we want to highlight that our codes were not carefully optimized to minimize runtime or to make optimal use of the hardware resources.

	\item \textit{Number of parameters in each model}: We provide these details in Sec.~\ref{subsec:plms} and Table~\ref{tab:odqa-ansgen}.

	\item \textit{Corresponding validation performance for each reported test result}: The re-ranking experiments does not require validation set for model selection, as we only perform inference for each query using the language model and retrieved passages. If the program committee or reviewers require the validation set performance, we will include it in the Appendix in the final version of the paper. Our ablations and analysis are conducted on the validation set of datasets. For the open-domain QA experiments, we also report the performance on the validation set.
	
	\item \textit{Explanation of evaluation metrics used, with links to code}: Our evaluation metrics are standard and widely used by the community. We provide their details in the main paper in Sec.~\ref{sec:exp-passage-retrieval}. The code is submitted with the paper.
\end{itemize}

\subsection{For all results involving multiple experiments, such as hyperparameter search}
\begin{itemize}
	\item \textit{The exact number of training and evaluation runs:} We provide training details for all models in Sec.~\ref{subsec:implementation_details}.

	\item \textit{Hyperparameter configurations for best-performing models}: We provide the hyperparameter settings in Appendix~\ref{appendix:train-hparams}.
	
	\item \textit{The bounds for each hyperparameter}: As described in Appendix~\ref{appendix:train-hparams}, our model and training setting uses standard hyperparameters such as different dropouts $\in [0, 1)$, warmup ratio of optimizer $\in [0.01, 0.05]$, weight regularization $\in [0, 1]$, and learning rate $\in [1e^{-4}, 1e^{-5}]$.
	
	\item \textit{Number of hyperparameter search trials}: maximum 5.
	
	\item \textit{The method of choosing hyperparameter values (e.g., uniform sampling, manual tuning, etc.) and the criterion used to select among them (e.g., accuracy)}: For the open-domain QA experiments, we performed manual hyperparameter tuning. We selected the best hyperparameter using EM results on the validation set.
	
	\item \textit{Summary statistics of the results (e.g. mean, variance, error bars, etc.)}: The re-ranking experiments are based on performing inference using open-source PLMs using a single prompt. As such, these summary statistics are not applicable to UPR.
	The open-domain QA experiments are compute expensive utilizing a lot of CPU and GPUs resources and take time in the range of tens of hours. Therefore, due to computational and time constraints performing multiple runs for each experiment was not feasible. Therefore, we adopted the approach of using the same seed value (1234) for all the training runs.
\end{itemize}

\subsection{For all datasets used}
\begin{itemize}
    \item \textit{Details of train/validation/test splits}: We use the standard training / dev / test splits whose details are provided in Sec.~\ref{subsec:datasets} and Table~\ref{tab:dataset_stats}.

	\item \textit{Relevant statistics such as number of examples and label distributions}: We provide dataset statistics details in Table~\ref{tab:dataset_stats}.

	\item \textit{An explanation of any data that were excluded, and all pre-processing steps}: We include the relevant details in Sec.~\ref{sec:exp-setup}.

	\item \textit{For natural language data, the name of the language(s)}: Our datasets are in English language.
	
	\item \textit{A zip file containing data or link to a downloadable version of the data}: All the datasets used in this work are open-source available and widely used by the community. Please refer to the respective dataset papers for the download links.
	
	\item \textit{For new data collected, a complete description of the data collection process, such as instructions to annotators and methods for quality control}: This is not applicable to this work.
\end{itemize}

\end{document}